%% file: main.tex
\documentclass[10pt,twocolumn,letterpaper]{article}

\usepackage[pagenumbers]{cvpr} 

\input{preamble}
\definecolor{cvprblue}{rgb}{0.21,0.49,0.74}
\usepackage[pagebackref,breaklinks,colorlinks,allcolors=cvprblue]{hyperref}
\usepackage{algorithm}

\usepackage{algorithmic}
\usepackage{multirow}
\usepackage{graphicx}
\usepackage{amsmath}
\usepackage{amssymb}
\usepackage{booktabs}
\usepackage{color}
\usepackage{colortbl}  
\usepackage{xcolor}
\usepackage[table]{xcolor}
\usepackage{array}
\usepackage{bbding}
\usepackage{tabularx}
\usepackage{bbding}
\usepackage{makecell}
\def\confName{CVPR}
\def\confYear{2026}

\title{Visual-Aware CoT: Achieving High-Fidelity Visual Consistency \\ in Unified Models}

\author{%
  Zixuan Ye\textsuperscript{1}\thanks{Work done during an internship at Kling Team, Kuaishou Technology. }\hspace{0.7em}
  Quande Liu\textsuperscript{2}\thanks{Corresponding author.}\hspace{0.7em}
  Cong Wei\textsuperscript{2}\hspace{0.7em}
  Yuanxing Zhang\textsuperscript{2}\hspace{0.7em}
  Xintao Wang\textsuperscript{2}\hspace{0.7em} \\
  Pengfei Wan\textsuperscript{2}\hspace{0.7em} 
  Kun Gai\textsuperscript{2}\hspace{0.7em}
  Wenhan Luo\textsuperscript{1}\footnotemark[2] \\ 
  \textsuperscript{1}The Hong Kong University of Science and Technology \\
  \textsuperscript{2}Kling Team, Kuaishou Technology\\
  \texttt{\href{https://zixuan-ye.github.io/VACoT}{https://zixuan-ye.github.io/VACoT}}
}

\begin{document}
\maketitle
\input{sec/0_abstract}    
\input{sec/1_intro}

\input{sec/2_Related}

\input{sec/3_Preliminary}

\input{sec/4_Method}

\input{sec/5_Exp}

\section{Conclusion}
\label{sec:conclusion}
In this work, we identify a key limitation in existing multimodal Chain-of-Thought reasoning: lack of visual context consistency in generation tasks. To address this challenge, we propose Visual-Aware CoT (VACoT), a framework that shifts from text-following to visually-aware reasoning in unified models. Our contributions include Adaptive Visual Planning to generate structured visual checklists for consistency preservation and Iterative Visual Correction for self-reflection and progressive refinement. Through two-stage training combining supervised fine-tuning and flow-GRPO with visual consistency rewards, we infuse visual awareness into unified models. Experiments show VACoT significantly outperforms existing approaches in multimodal generation with improved visual context consistency.

{
    \small
    \bibliographystyle{ieeenat_fullname}
    \bibliography{main}
}

\input{sec/X_suppl}

\end{document}

%% file: sec/0_abstract.tex
\begin{abstract}
Recently, the introduction of Chain-of-Thought (CoT) has largely improved the generation ability of unified models. However, it is observed that the current thinking process during generation mainly focuses on the text consistency with the text prompt, ignoring the \textbf{visual context consistency} with the visual reference images during the multi-modal generation, e.g., multi-reference generation. The lack of such consistency results in the failure in maintaining key visual features (like human ID, object attribute, style). To this end, we integrate the visual context consistency into the reasoning of unified models, explicitly motivating the model to sustain such consistency by 1) Adaptive Visual Planning: generating structured visual check list to figure out the visual element of needed consistency keeping, and 2) Iterative Visual Correction: performing self-reflection with the guidance of check lists and refining the generated result in an iterative manner. To achieve this, we use supervised finetuning to teach the model how to plan the visual checking, conduct self-reflection and self-refinement, and use flow-GRPO to further enhance the visual consistency through a customized visual checking reward. The experiments show that our method outperforms both zero-shot unified models and those with text CoTs in multi-modal generation, demonstrating higher visual context consistency.

\end{abstract}

%% file: sec/1_intro.tex
\section{Introduction}
\label{sec:intro}

\begin{figure*}
    \centering
    \includegraphics[width=1\linewidth]{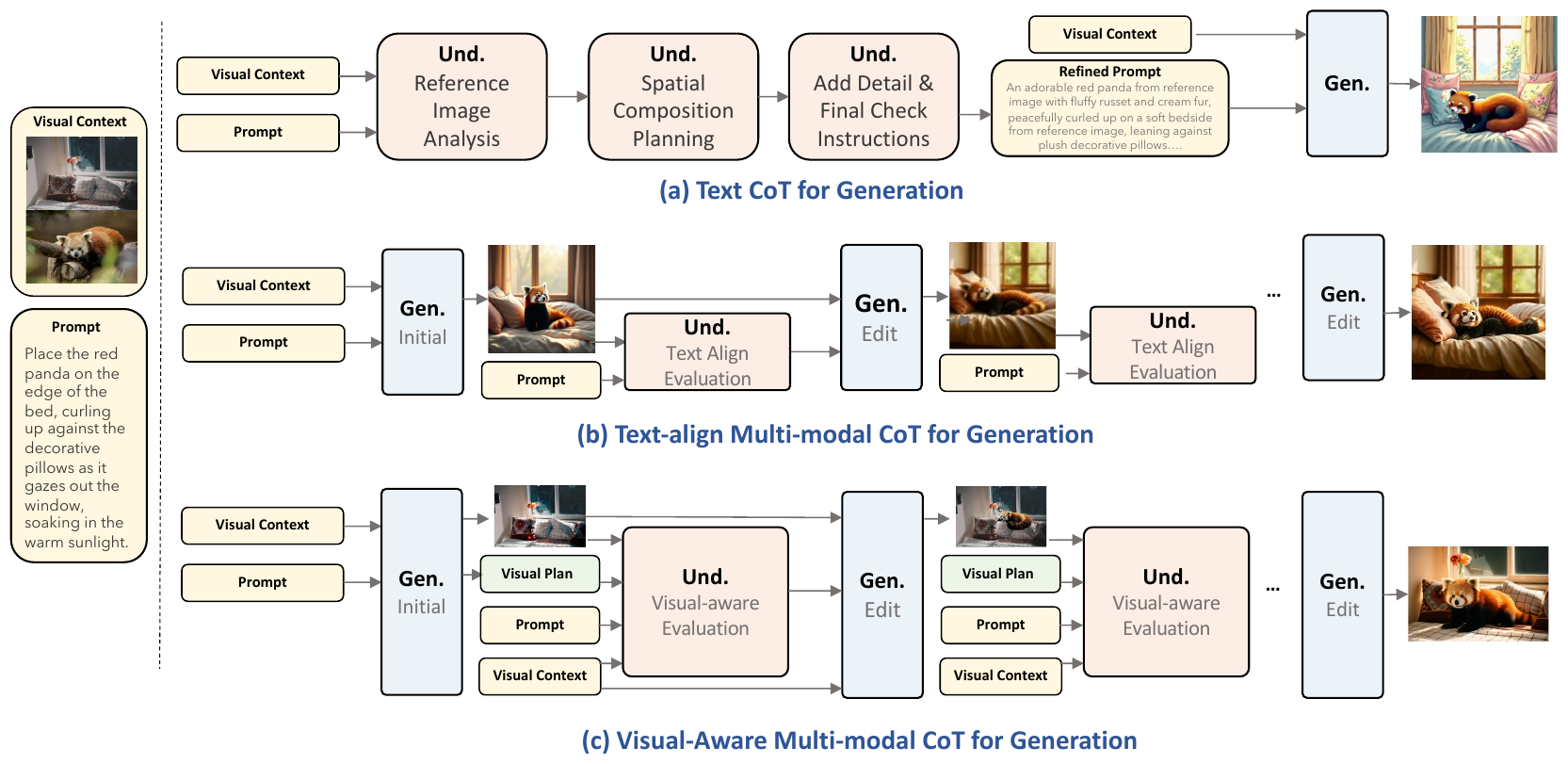}
    \caption{\textbf{Comparison of Chain-of-Thought approaches for  generation.} (a) Text CoT uses text-only understanding modules for analysis and planning. (b) Text-align Multi-modal CoT iteratively refines generation through text-alignment evaluation. (c) Visual-Aware Multi-modal CoT (Ours) incorporates visual planning and visual-aware evaluation for improved visual consistency.}
    \label{fig:teaser}
\end{figure*}
The emergence of unified models such as GPT-4o~\cite{openai_gpt4o_2025} and UniWorld~\cite{lin2025uniworld} has shown remarkable capabilities in handling both understanding and generation within one model. Drawing inspiration from the success of Chain-of-Thought (CoT) reasoning, which shows effectiveness in enhancing output quality through thinking-with-image approaches for understanding~\cite{su2025thinking} and prompt reasoning for generation, recent works including Uni-CoT~\cite{qin2025uni} and UiG~\cite{lyu2025understanding} explore the integration of multimodal CoT reasoning into unified models. This integration facilitates self-reflection mechanisms to iteratively refine the result, leading to significant improvements in final generation quality.

However, we find a critical limitation in existing CoT approaches, i.e., they mainly focus on text consistency, ensuring text alignment with input prompts during text-to-image (T2I) generation, while ignoring \textbf{visual context consistency} with input visual contexts during multi-modal generation. As shown in Fig.~\ref{fig:teaser}, the lack of visual consistency can lead to sub-optimal generation results in tasks requiring complex visual information fusion, such as multi-reference image generation or editing. Specifically, it causes significant discrepancies between generated content and input visual contexts, and results in the failure to preserve fine-grained visual features, including human identity, object attributes, and stylistic elements. \textit{Therefore, we desire a paradigm shift from ``text-following'' to ``visually-aware'' reasoning in generation-understanding unified models.}

To achieve this, we propose VACoT (Visual-Aware CoT), a novel framework that integrates visual context consistency into the reasoning process of unified models. Our approach explicitly motivates models to maintain visual consistency through two key innovations. First, \textbf{Adaptive Visual Planning} generates structured visual checklists that systematically identify visual elements requiring consistency preservation, enabling models to explicitly reason about which visual features should be maintained. Second, \textbf{Iterative Visual Correction} builds upon the visual checklists to perform self-reflection and iterative refinement, where the model evaluates its output under the guidance of the visual checking list and progressively improves generation quality.

To equip the model with such visual-aware abilities, we propose a two-stage training strategy to infuse the visual awareness ability into the unified model. In the first stage, we construct visual planning and self-reflection datasets to develop robust reasoning patterns through supervised fine-tuning. In the second stage, we employ flow-GRPO~\cite{liu2025flow} with a carefully designed visual consistency reward function that quantifies the alignment between generated outputs and input visual contexts.

Extensive experiments demonstrate that our method significantly outperforms both zero-shot unified models and those equipped with traditional text-based CoT reasoning in multimodal generation tasks. The results show substantial improvements in visual context consistency while maintaining competitive performance in text alignment. The main contributions of this work are:
\begin{itemize}
    \item We propose VACoT, a novel framework that integrates adaptive visual planning and iterative visual correction into unified model reasoning, effectively enhancing visual consistency during multimodal generation while addressing the gap in existing text-focused CoT mechanisms. 

    \item We develop a two-stage training strategy that injects visual-aware capabilities into unified models through supervised fine-tuning on carefully constructed visual planning and self-reflection datasets, followed by flow-GRPO with a novel visual consistency reward function that quantifies alignment between generated outputs and input visual contexts.
    \item Our method achieves state-of-the-art performance on the multi-reference generation benchmark, while also preserving strong capabilities for fundamental text-to-image generation tasks.

\end{itemize}

%% file: sec/2_Related.tex
\section{Related Work}
\label{sec:related}

\noindent\textbf{Unified Models.}
With remarkable advancements in both \textbf{generative models} and \textbf{understanding capabilities} of multimodal models, recent work aims to unify these tasks within single frameworks~\cite{deng2025emerging, team2024chameleon, lin2025uniworld, pan2025transfer}. Some researchers try to use autoregressive frameworks to handle both tasks. Works like Transfusion~\cite{zhou2024transfusion}, Chameleon~\cite{team2024chameleon}, Emu3~\cite{wang2024emu3}, and Show-o~\cite{xie2024show} convert images into discrete visual tokens using tokenizers. However, the extracted image tokens cannot meet the high-level semantic requirement of understanding and rich details for generation. Therefore, another stream identifies that understanding and generation have different requirements and uses separate encoders for each task, focusing on bridging VLMs~\cite{driess2023palm,peng2023kosmos,zhu2023minigpt,wang2024qwen2,bai2025qwen2,chen2024internvl,li2023blip,lin2023video,gao2023llama,zhang2023llama,wei2025univideo} and diffusion models. For example, UniWorld~\cite{lin2025uniworld} extracts high-level semantics from VLMs as generation conditions, while MetaQueries~\cite{pan2025transfer} introduces learnable query tokens as an interface between frozen MLLMs and diffusion transformers. Alternatively, Janus~\cite{chen2025janus} employs separate encoders for understanding and generation to address task conflicts, while BAGEL~\cite{deng2025emerging} uses a Mixture-of-Transformer-Experts architecture under a unified decoder-only transformer. Our work builds upon these unified models to enhance generation controllability, particularly for maintaining visual consistency with input images.

\begin{figure*}[!t]
    \centering
    \includegraphics[width=1\linewidth]{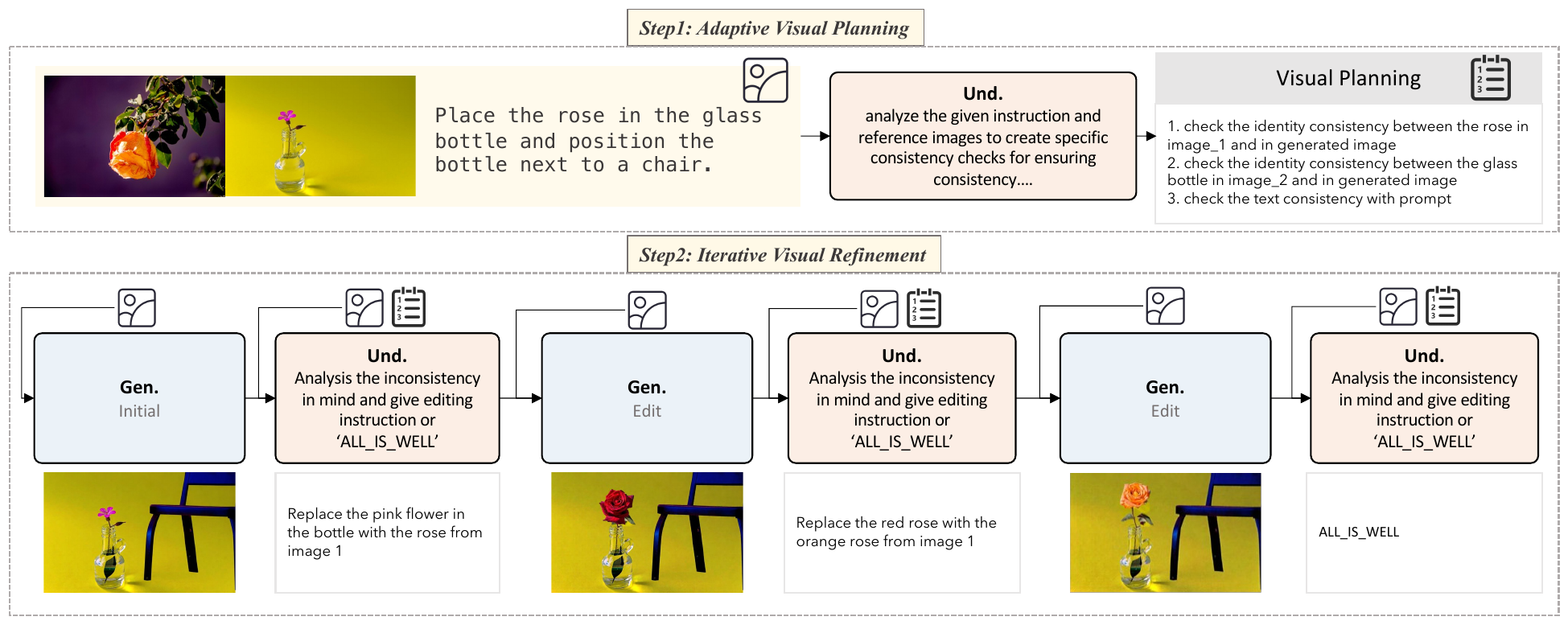}
    \caption{\textbf{Adaptive Visual Planning and Iterative Visual Correction Process of Our Method.}}
    \label{fig:iterative}
\end{figure*}

\noindent\textbf{Reasoning in Generation.}
Since the breakthrough of CoT reasoning~\cite{wei2022chain}, this paradigm has been widely adopted across various domains to decompose complex tasks into structured intermediate steps. In large language models, CoT enables step-by-step problem solving~\cite{xia2024beyond,zhang2024chain,deng2024explicit}. For multimodal understanding, it facilitates ``thinking with images'' approaches, such as first locating target regions before answering questions~\cite{wang2025multimodal,jiang2025t2i,xu2024llava}. In text-to-image generation, CoT has been applied through candidate selection, e.g., ImageCoT~\cite{zhang2025let} generates multiple candidates and applies CoT reasoning for final selection; or through step-by-step prompt refinement for detailed layout planning~\cite{jiang2025t2i,zhang2025reasongen}. With the emergence of unified understanding-generation models, multimodal CoT becomes feasible. For instance, Uni-CoT~\cite{qin2025uni} and UiG~\cite{lyu2025understanding} employ the understanding capability to evaluate the alignment between generated images and text prompts, using this feedback to iteratively guide and refine the generation process. However, these methods focus on text-level planning and verification, addressing \textbf{does the generation align with the text prompt?} but lack visual-aware self-reflection to answer \textbf{does the generation align with the input images?} This limitation becomes evident in tasks requiring complex visual information fusion, such as multi-reference image generation, style transfer, or content-aware editing, where maintaining visual consistency and fidelity remains a significant challenge.

%% file: sec/3_Preliminary.tex
\section{Preliminary: Unified Model}
\label{sec:preliminary}

To enable both planning and self-refinement within one model, we prefer a unified understanding and generation model to support multi-modal chain-of-thought. 

\noindent\textbf{Architecture.}
We choose BAGEL~\cite{deng2025emerging} as our base model. BAGEL uses a unified decoder-only Transformer with Mixture-of-Transformer-Experts design, using two experts for understanding and generation respectively. Both experts share multi-modal token sequences through unified self-attention, enabling flexible cross-modal fusion. The understanding expert uses ViT for image tokens and a two-layer connector for LLM feature alignment, while the generation expert employs a pretrained VAE for encoding/decoding between hidden states and pixel space.

\noindent\textbf{Training.}
The training of the unified model consists of two parts: cross-entropy (CE) loss for text prediction and mean squared error (MSE) loss computed for the velocity prediction in the flow matching process.

\noindent\textbf{Inference.}
With the unified architecture, BAGEL can flexibly handle interleaved multi-modal sequences. During inference, the model generates interleaved multi-modal content by appending the generation result of each step to the sequence. This capability establishes the foundation for our model to perform adaptive visual consistency planning and iterative correction of generated results through multi-modal chain-of-thought reasoning.

%% file: sec/4_Method.tex
\begin{figure*}
    \centering
    \includegraphics[width=1\linewidth]{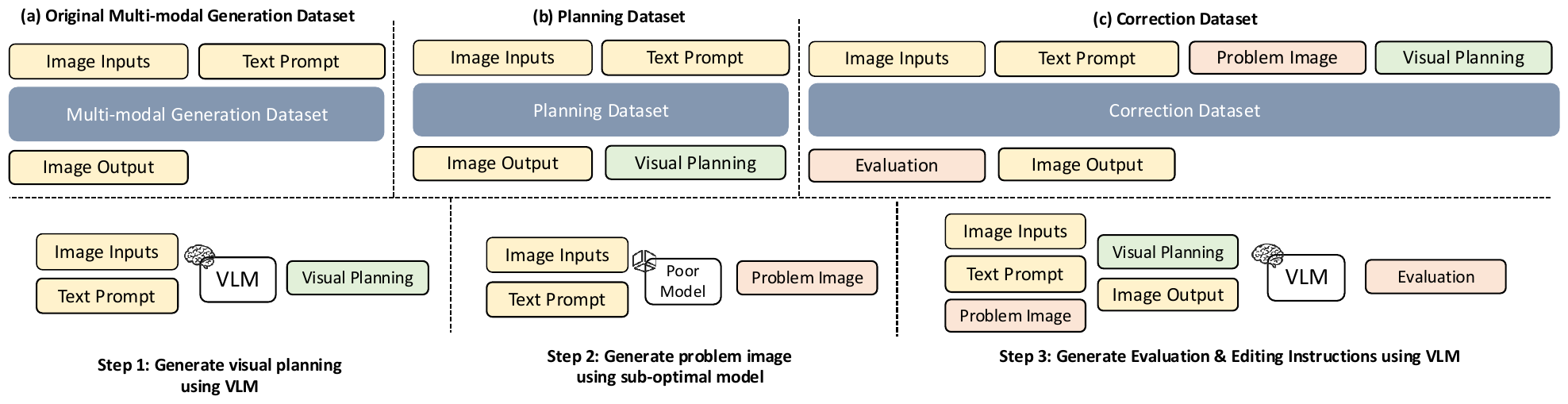}
    \caption{\textbf{Dataset construction for the planning and correction process.}}
    \label{fig:dataset}
\end{figure*}

\section{Visual-Aware CoT}
\label{sec:vacot}

In this section, we propose VACoT (Visual-Aware Chain-of-Thought), a novel framework that integrates visual context consistency into the reasoning process of unified generation-understanding models. We achieve this target through two key components: Adaptive Visual Planning (Sec~\ref{sec:planning}), responsible for explicitly infer the checking lists; and Iterative Visual Correction (Sec~\ref{sec:correction}), progressively evaluate and refine the generated results under the guidance of checking lists. The visual consistency GRPO (Sec~\ref{sec:grpo}) is designed to further enhance the visual consistency.

\begin{figure*}
    \centering
    \includegraphics[width=1\linewidth]{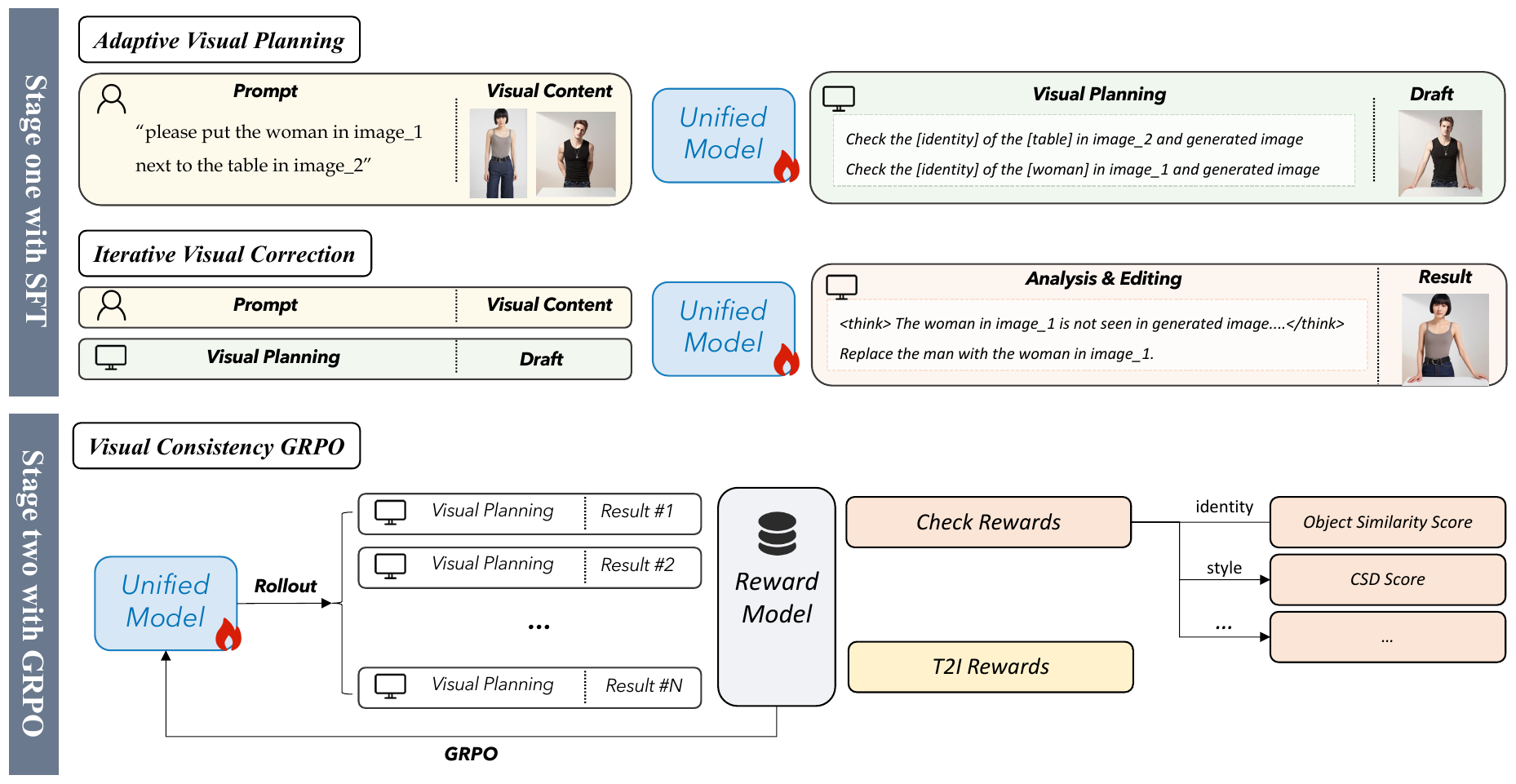}
    \caption{\textbf{Our training pipeline of Visual-Aware CoT.}}
    \label{fig:pipeline}
\end{figure*}

\subsection{Problem Definition}
\label{sec:problem}

Given the text prompt $T$ and the visual context $V = \{v_1, v_2, \ldots, v_n\}$ containing $n$ multiple reference images as the model input, our goal is to generate output image $Y$ through a $\mathcal{T}$ steps denoising process that maintains both textual alignment with $T$ and visual consistency with $V$.
Denote the corresponding reverse time trajectories as $\{x_\mathcal{T}, x_{\mathcal{T}-1}, \ldots, x_0=Y\}$.

While previous method have tried to enhance the textual alignment with $T$ using CoT, they overlook the visual consistency with $V$ during multi-modal generation with reference images as input, which is also a deciding part of the quality of the generated image. To enable a visual-aware CoT, we propose to develop an explicit reasoning approach to identify what visual elements need consistency, evaluate the consistency, and correct the generation results. To achieve this, we propose an iterative inference process that consists of \textbf{Adaptive Visual Planning} and \textbf{Iterative Visual Correction}. As illustrated in Algorithm~\ref{alg:inference} and Fig.~\ref{fig:iterative}, it begins with visual planning and initial generation given the user inputs, then iteratively refine the generated output through self-evaluation until it reaches satisfactory decision in evaluation or it reaches the maximum iterations $N$.

\begin{algorithm}[t]
\caption{VACoT Iterative Inference}
\label{alg:inference}
\begin{algorithmic}[1]
\STATE \textbf{Input:} Text $T$, Visual context $V$, Max iterations $N$, The unified model $\pi_\theta$ with parameter $\theta$.
\STATE \textbf{Step 1:} $(Z_{\text{plan}}, Y_0) \leftarrow \pi_\theta(T, V)$ \COMMENT{Planning \& Initial generation}
\STATE {\color{gray}// $Z_{\text{plan}}$: planning check list, $Y_0$: initial image}
\STATE $Y_{\text{current}} \leftarrow Y_0$ 
\FOR{$i = 1$ \textbf{ to } $N$}
    \STATE $(Z_{\text{eval}}, Y_{\text{next}}) \leftarrow \pi_\theta(T, V, Z_{\text{plan}}, Y_{\text{current}})$ \COMMENT{Evaluation \& Refinement}
    \STATE {\color{gray}// $Z_{\text{eval}}$: evaluation and editing instructions}
    \STATE{\color{gray}//$Y_{\text{next}}$: refined image}
    \STATE $Y_{\text{current}} \leftarrow Y_{\text{next}}$
    \IF{$Z_{\text{eval}}$ indicates satisfaction}
        \STATE \textbf{break}
    \ENDIF
\ENDFOR
\STATE \textbf{Return:} $Y_{\text{final}} \leftarrow Y_{\text{current}}$ {\color{gray}// $Y_\text{final}$: Final image}
\end{algorithmic}
\end{algorithm}

\subsection{Adaptive Visual Planning}
\label{sec:planning}

To facilitate high-quality evaluation in subsequent steps, we propose to generate an explicit checklist during the planning stage. The planning checklist $Z_{\text{plan}}$ represents a structured reasoning chain that systematically identifies what visual consistency requirements need to be verified between the input context and generated output. Drawing inspiration from thinking-in-image methods~\cite{jiang2025t2i,wang2025multimodal}, which demonstrate the effectiveness of first localizing relevant regions and then providing the answer, our method similarly first determines the checking targets through planning, then conducts systematic evaluation. The planning checklist $Z_{\text{plan}}$ decomposes complex visual consistency requirements into manageable components that align with human reasoning processes. This structured approach adaptively clarifies what types of consistency need to be checked and identifies the specific checking targets. The checking types are categorized by:
\begin{itemize}[leftmargin=*, itemsep=2pt]
    \item \textbf{Identity:} Identity of the characters, judging the preservation from the input visual context (e.g., facial features)
    \item \textbf{Style:} Aesthetic style of the image, checking the maintenance from the input visual reference (e.g., artistic style, color palette, texture patterns)  
    \item \textbf{Attribute:} Attribute preservation, verifying the consistency of specific visual properties (e.g., color, shape, size, spatial relationships)
\end{itemize}
Each checklist item $z_i \in Z_{\text{plan}}$ is defined as:
\begin{equation}
z_i = \{\texttt{check\_type}, \texttt{source}, \texttt{target}\}
\end{equation}
where \texttt{source} and \texttt{target} specify which visual elements from the input context should be preserved in which parts of the generated output. This structured approach enables the model to systematically decompose complex visual consistency requirements into manageable, verifiable components that adapt to user input. For example, given the text prompt $T$ = ``the woman in \texttt{image\_1} is dancing, in the artistic style of \texttt{image\_2}'', the checklist would include: (1) checking identity consistency of the woman between \texttt{image\_1} and the generated image, and (2) verifying style consistency between \texttt{image\_2} and the generated image.

\noindent\textbf{Dataset Construction.}
To enable the learning of this ability, we construct a dataset $D_{\text{planning}}$ containing data tuples $(T, V, Z_{\text{plan\_GT}}, Y_{\text{final\_GT}})$, where:
\begin{itemize}
    \item $(T, V)$ represents the original input for our task (e.g., multi-reference generation).
    \item $Y_{\text{final\_GT}}$ is the high-quality ground-truth image.
    \item $Z_{\text{plan\_GT}}$ is automatically generated by a powerful VLM.
\end{itemize}
As illustrated in Figure~\ref{fig:dataset}, we leverage existing multi-modal generation datasets and augment them with planning annotations. Given existing datasets that contain $(T, V, Y_{\text{final\_GT}})$ tuples from tasks such as multi-reference image generation, style transfer, and identity-preserving generation, we only need to generate the additional $Z_{\text{plan\_GT}}$ component. Specifically, we use 4k multi-reference generation data from Echo-4o~\cite{ye2025echo} as our base dataset. We employ Gemini~\cite{google_gemini2_flash} to analyze the text prompt $T$ and input images $V$, then automatically generate structured checklists that specify what consistency checks should be performed. After comparing with other VLM models, we found that Gemini produces higher quality reasoning outputs and maintains more stable output format structures, making it our preferred choice for checklist generation.

\begin{figure*}
    \centering
    \includegraphics[width=1\linewidth]{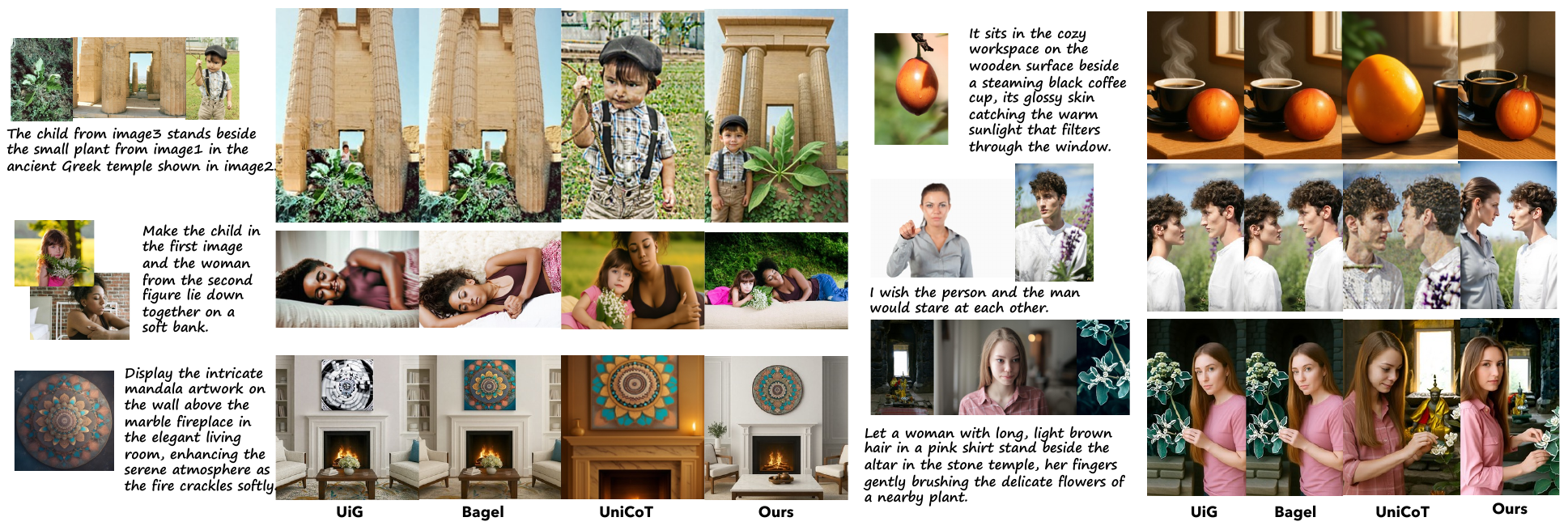}
    \caption{\textbf{Qualitative Comparison on Multi-Reference Generation}. Our VACoT demonstrates superior identity consistency and visual coherence compared to baseline methods. BAGEL often fails to maintain character consistency across references, while UiG and UniCoT focus on text alignment rather than visual consistency. Our approach effectively preserves identity features and generates more accurate multi-reference images.}
    \label{fig:maincomp}
\end{figure*}

\subsection{Iterative Visual Correction}
\label{sec:correction}

Building upon the visual checklist, our Iterative Visual Correction mechanism enables the model to perform self-reflection and progressive refinement. The model evaluates its current generation under the guidance of the visual checklist $Z_{\text{plan}}$ and produces editing instructions $Z_{\text{eval}}$. The $Z_{\text{eval}}$ is then added to the context sequence, guiding the model to edit the generated image toward the target direction.

Specifically, given the current generated image $Y_{\text{current}}$ and the planning checklist $Z_{\text{plan}}$, the model first performs consistency evaluation as
\begin{equation}
Z_{\text{eval}} = f_{\text{evaluate}}(Y_{\text{current}}, Z_{\text{plan}}, V, T),
\end{equation}
where $Z_{\text{eval}}$ contains detailed feedback on which consistency requirements are satisfied and provides editing instructions. $Z_{\text{eval}}$ is then incorporated into the generation context to guide the correction process as
\begin{equation}
Y_{\text{corrected}} = f_{\text{edit}}(T, V, Z_{\text{plan}}, Y_{\text{current}}, Z_{\text{eval}}).
\end{equation}

As shown in Algorithm~\ref{alg:inference}, this iterative process continues until the consistency requirements are met or a maximum number of iterations is reached.

\noindent\textbf{Dataset Construction.}
As shown in Fig.~\ref{fig:dataset}, building upon $D_{\text{planning}}$, we construct the correction dataset $D_{\text{correction}}$ by augmenting each tuple with additional components for the correction learning. Specifically, for each $(T, V, Z_{\text{plan\_GT}}, Y_{\text{final\_GT}})$ in $D_{\text{planning}}$, we generate

\begin{itemize}
    \item $Y_{\text{negative}}$: Lower-quality images that violate consistency requirements, obtained by using a weaker baseline model.
    \item $Z_{\text{eval\_GT}}$: Ground-truth evaluation feedback generated by the VLM, which analyzes $Y_{\text{negative}}$ against $Z_{\text{plan\_GT}}$ and provides specific editing instructions.
\end{itemize}

To ensure the diversity of these visual queries, we adjust the parameters and CFG scale of the original BAGEL to generate $Y_{\text{negative}}$ samples with varied failure modes. Our manual inspection confirmed diverse failure cases including identity loss, identity inconsistency, and text-image misalignment. This results in $D_{\text{correction}}$ containing tuples $(T, V, Z_{\text{plan\_GT}}, Y_{\text{negative}}, Z_{\text{eval\_GT}}, Y_{\text{final\_GT}})$, where the model learns to evaluate inconsistencies and generate correction instructions that transform $Y_{\text{negative}}$ into $Y_{\text{final\_GT}}$.

\noindent\textbf{Training Strategy.}
As shown in Fig.~\ref{fig:pipeline}, at the first stage, we train our model using supervised fine-tuning (SFT) on the combined datasets $D_{\text{planning}}$ and $D_{\text{correction}}$. The datasets are mixed during training, allowing the model to jointly learn both planning and correction capabilities.

\subsection{Visual Consistency GRPO}
\label{sec:grpo}
To optimize visual reasoning and consistency quality, we employ a reinforcement learning framework using flow-GRPO~\cite{liu2025flow} with designed visual consistency rewards. The composite reward function evaluates the final generation by measuring consistency with both text and visual contexts:
\begin{equation}
R_{\text{total}}(Y_{\text{final}},V,T) = R_{\text{visual}}(Y_{\text{final}}, Z_{\text{plan}}, V) + R_{\text{text}}(Y_{\text{final}},T).
\end{equation}
The visual consistency reward $R_{\text{visual}}$ dynamically evaluates each checklist item using type-specific visual similarity metrics. For identity preservation, we use GroundingDINO~\cite{liu2024grounding} to locate the target object, then compute DINO similarity~\cite{caron2021emerging} as the score; for style consistency, we employ CSD-Score~\cite{somepalli2024measuring}. The text consistency reward $R_{\text{text}}$ uses CLIP score to measure alignment between $x_0=Y_{\text{final}}$ and $T$. The advantage of the $i$-th sample (yielded $\{x_\mathcal{T}^i,\ldots, x_0^i$\}) in a group of size $G$ is calculated as:
\begin{equation}
\hat{A}_t^i = \frac{R(x_0^i,V,T) - \mathrm{mean}(\{R(x_0^j,V,T)\}_{j=1}^G)}{\mathrm{std}(\{R(x_0^j,V,T)\}_{j=1}^G)}\,.
\end{equation}
We optimize the policy model by maximizing the objective:
\begin{equation}
\mathcal{J}_{\text{Flow-GRPO}}(\theta) = \mathbb{E}_{V,T, \{x_0^i\}^G_{i=1}\sim\pi_{\theta_\text{old}}} \Big[ f(r, \hat{A}, \theta, \epsilon, \beta) \Big],
\end{equation}
where $\epsilon$ and $\beta$ are hyperparameters, and
\begin{equation}
\begin{split}
&f(r, \hat{A}, \theta, \epsilon, \beta) = \frac{1}{G}\sum_{i=1}^G \frac{1}{\mathcal{T}} \sum_{t=1}^{\mathcal{T}} 
\min\big(r_t^i(\theta) \hat{A}_t^i, \\
&\quad \text{clip}(r_t^i(\theta),1-\epsilon,1+\epsilon)\hat{A}_t^i \big) - \beta D_{\text{KL}}(\pi_\theta || \pi_{\text{ref}})
\end{split}
\end{equation}
with $r_t^i(\theta) = \frac{\pi_\theta(x_{t-1}^i|x_t^i,V,T)}{\pi_{\theta_{\text{old}}}(x_{t-1}^i|x_t^i,V,T)}$.

%% file: sec/5_Exp.tex
\begin{table*}[h!]
\centering
\footnotesize
\addtolength{\tabcolsep}{1pt}
\caption{\textbf{Evaluation of text-to-image generation ability on GenEval benchmark.} `Gen. Only` stands for an image
generation model, and ‘Unified’ denotes a model that has both understanding and generation capabilities.}
\begin{tabular}{llcccccccc}
\toprule
\textbf{Type} & \textbf{Model} & \textbf{Single Obj.} & \textbf{Two Obj.} & \textbf{Counting} & \textbf{Colors} & \textbf{Position} & \textbf{Color Attri.} & \textbf{Overall↑} \\
\midrule
\multirow{7}{*}{\textbf{Gen. Only}} 
& Emu3-Gen \cite{wang2024emu3} & 0.98 & 0.71 & 0.34 & 0.81 & 0.17 & 0.21 & 0.54 \\
& SDXL \cite{podell2023sdxl} & 0.98 & 0.74 & 0.39 & 0.85 & 0.15 & 0.23 & 0.55 \\
& DALL-E 3 \cite{betker2023improving} & 0.96 & 0.87 & 0.47 & 0.83 & 0.43 & 0.45 & 0.67 \\
& SD3-Medium \cite{esser2024scaling} & 0.99 & 0.94 & 0.72 & 0.89 & 0.33 & 0.60 & 0.74 \\
& FLUX.1-dev \cite{blackforestlabs_flux_2024} & 0.98 & 0.93 & 0.75 & \textbf{0.93} & 0.68 & 0.65 & 0.82 \\
\midrule
\multirow{9}{*}{\textbf{Unified}} 
& SEED-X \cite{ge2024seed} & 0.97 & 0.58 & 0.26 & 0.80 & 0.19 & 0.14 & 0.49 \\
& TokenFlow-XL \cite{qu2025tokenflow} & 0.95 & 0.60 & 0.41 & 0.81 & 0.16 & 0.24 & 0.55 \\
& Show-o \cite{xie2024show} & 0.98 & 0.80 & 0.66 & 0.84 & 0.31 & 0.50 & 0.68 \\
& Janus-Pro-7B \cite{chen2025janus} & 0.99 & 0.89 & 0.59 & 0.90 & \textbf{0.79} & 0.66 & 0.80 \\
& MetaQuery-XL \cite{pan2025transfer} & - & - & - & - & - & - & 0.80 \\
& BAGEL \cite{deng2025emerging} & 0.99 & 0.92 & 0.78 & 0.87 & 0.53 & 0.64 & 0.79 \\
& UiG \cite{lyu2025understanding} & 0.99 & 0.92 & 0.81 & 0.89 & 0.61 & 0.69 & 0.82 \\
& Uni-CoT \cite{qin2025uni} & 0.99 & 0.95 & \textbf{0.82} & 0.89 & 0.60 & \textbf{0.72} & 0.83 \\
\rowcolor{gray!15}
& \textbf{Ours} &  \textbf{0.99} & \textbf{0.95} & 0.80& 0.90 & 	0.66& 0.71	& \textbf{0.84} \\
\bottomrule
\end{tabular}

\label{tab:geneval}
\end{table*}

\section{Experiments}
\label{sec:experiments}

In this section, we provide the evaluation of Visual-Aware CoT, validating its effectiveness across various generation tasks, including instruction-following image generation and multi-reference image generation. Additionally, we perform ablation studies to assess the impact of different training stages and demonstrate the importance of adaptive visual planning and iterative visual refinement components.

\subsection{Benchmark}
We choose multi-reference generation as our main focus task since it is the most representative task in multi-modal generation. We evaluate the performance on OmniContext~\cite{wu2025omnigen2}, a benchmark for multi-reference generation, compared with our baseline model BAGEL~\cite{deng2025emerging} and other text-alignment CoT methods, including UiG~\cite{lyu2025understanding} and Uni-CoT~\cite{qin2025uni}. Additionally, to validate whether it affects the T2I generation performance, we evaluate on GenEval~\cite{ghosh2023geneval}, a comprehensive benchmark for the T2I generation.

\subsection{Main Results}

\begin{table*}[t!]
\small
\addtolength{\tabcolsep}{3pt}
\caption{\textbf{Performance Comparison on OmniContext}. The table shows performance scores across different evaluation metrics for various models on MULTIPLE and SCENE datasets.}
\begin{tabular}{l|ccc|ccc|c}
\toprule
\multirow{2}{*}{\textbf{Method}} & \multicolumn{3}{c|}{\textbf{MULTIPLE}} & \multicolumn{3}{c|}{\textbf{SCENE}} & \multirow{2}{*}{\textbf{Average↑}} \\
\cmidrule{2-7}
& \textbf{Character} & \textbf{Object} & \textbf{Char. + Obj.} & \textbf{Character} & \textbf{Object} & \textbf{Char. + Obj.} & \\
\midrule
Gemini-2.0-flash \citep{google_gemini2_flash} & 2.91 & 2.16 & 3.80 & 3.02 & 3.89 & 2.92 & 3.12 \\
GPT-4o \cite{openai_gpt4o_2025} & \textbf{9.07} & 8.95 & \textbf{8.54} & \textbf{8.90} & 8.44 & \textbf{8.60} & \textbf{8.75} \\
\midrule
UNO \cite{wu2025less} & 2.54 & 6.51 & 4.39 & 2.06 & 4.33 & 4.37 & 4.03 \\
BAGEL \cite{deng2025emerging} & 5.17 & 6.64 & 6.24 & 4.07 & 5.71 & 5.47 & 5.55 \\
OmniGen \cite{xiao2025omnigen} & 5.65 & 5.44 & 4.68 & 3.59 & 4.32 & 5.12 & 4.8 \\
OmniGen2 \cite{wu2025omnigen2} & 7.11 & 7.13 & 7.45 & 6.38 & 6.71 & 7.04 & 6.97 \\
Echo-4o \cite{ye2025echo} & 8.07 &  7.50 & 8.29 & 8.62 &  8.00 &  8.08 &  8.09 \\
\midrule
UiG \cite{lyu2025understanding}& 5.47& 8.26& 8.22& 5.05& 7.06& 7.08& 6.85\\
Uni-CoT \cite{qin2025uni} & 7.12& 8.84& 7.97& 7.07& 8.16& 8.20& 7.89\\
\rowcolor{gray!20}
\textbf{Ours} & 7.82&  \textbf{9.21}& 8.30& 7.55&  \textbf{8.67}&  7.99&  8.26\\
\bottomrule
\end{tabular}
\label{tab:omnicontext}
\end{table*}

\noindent\textbf{Multi-Reference Generation Results.}
The quantitative results on the OmniContext benchmark are presented in Table~\ref{tab:omnicontext}. Our VACoT achieves outstanding performance across all evaluation metrics on both MULTIPLE and SCENE datasets. Compared to the baseline BAGEL model, our approach demonstrates significant improvements in character consistency, object preservation, and combined character-object generation tasks. Notably, in the multiple object setting and scene-with-object setting, our method even surpasses the performance of GPT-4o. When compared with other CoT methods applied to BAGEL, our approach consistently outperforms UiG and Uni-CoT, as these methods focus only on text alignment without considering visual consistency. \textbf{Detailed score of each category can be found in Supplementary Material.}

The qualitative comparison on OmniContext~\cite{wu2025omnigen2} is shown in Fig.~\ref{fig:maincomp}. Our method exhibits more stable and outstanding identity consistency compared to baselines. BAGEL often misses identities or fails to maintain visual consistency with reference images, while UiG and UniCoT, as text-based CoT methods, focus only on content existence rather than visual consistency, thus failing to identify identity-related errors as clearly shown in the middle-right case. In contrast, our method focuses on visual context consistency, enabling better identity preservation and more accurate multi-reference generation.

\noindent\textbf{Text-to-Image Generation Results.}
To ensure that our visual-aware reasoning training does not compromise general T2I generation capabilities, we evaluate our method on the GenEval benchmark. Without the visual context, we use a fixed template to guide the self-reflection, i.e., check the consistency between the user prompt and the generated image. As shown in Table~\ref{tab:geneval}, our method shows competitive performance across various compositional generation tasks, and achieves the highest overall score compared with both baseline and other CoT methods. Contrary to the common expectation that additional training on specific scenario might degrade baseline performance, our Visual-Aware CoT method not only preserves the strong T2I generation capabilities while enhancing visual consistency, but also demonstrates unexpected improvements in several compositional generation metrics. This suggests that the visual-aware ability training implicitly encourages more coherent and structured image generation, enhancing visual consistency without sacrificing, and in some cases even boosting, the fundamental text-to-image performance.

\begin{figure}
    \centering
    \includegraphics[width=1\linewidth]{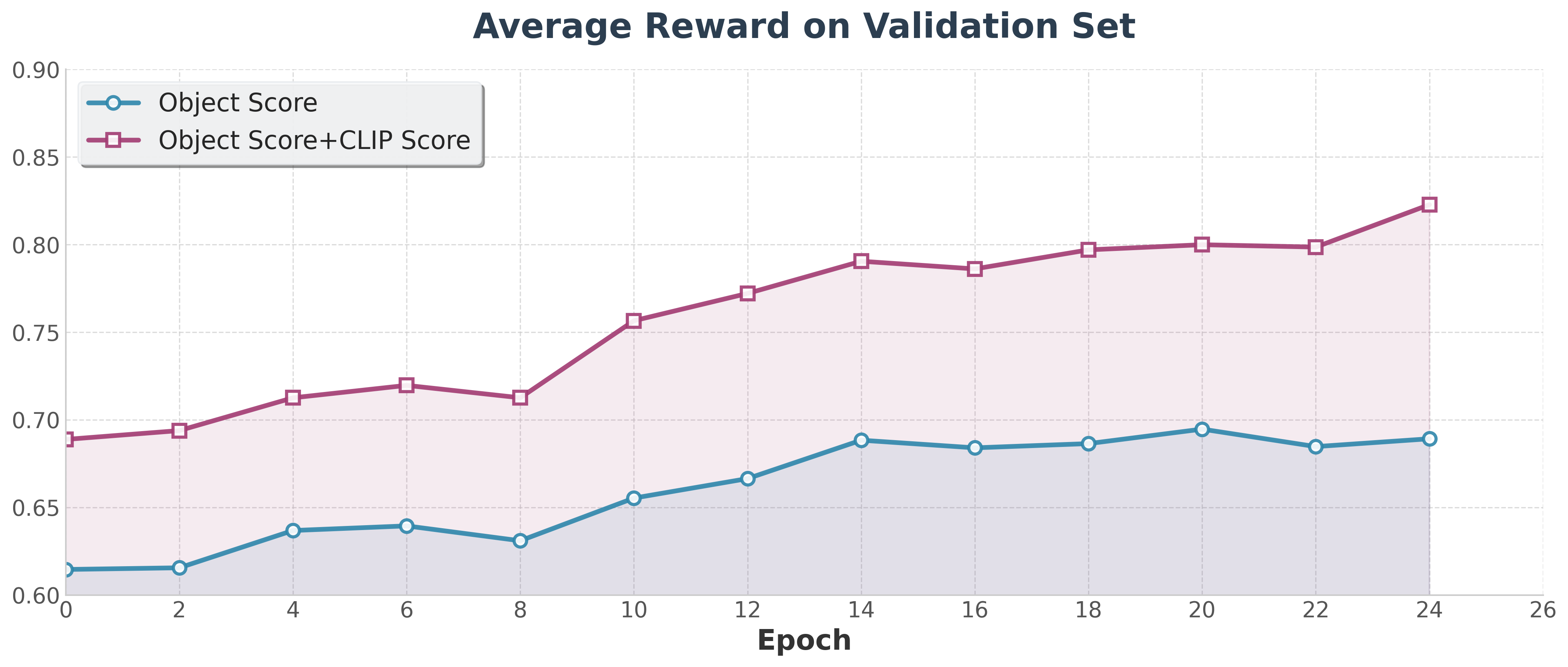}
    \caption{\textbf{The reward score on validation set in GRPO with different reward settings.}}
    \label{fig:reward}
\end{figure}

\subsection{Ablation Studies}


\noindent\textbf{Impact of Training Stages.}
To validate the specific effect of the two-stage training, we conduct an ablation study on the training stages, with results shown in Table~\ref{tab:ablation_main}. Compared with direct generation (BAGEL gen-only mode, L1), even with the original BAGEL understanding to plan and evaluate (L2), our planning-then-checking process designed in VACoT significantly improves generation quality. Furthermore, both SFT and GRPO stages contribute additional gains, and the complete two-stage training pipeline maximizes the overall effectiveness. The reward score on the validation set is shown in Fig.~\ref{fig:reward}, demonstrating that both reward settings can steadily improve model performance throughout the training process.

\begin{table}[t!]
\caption{\textbf{Ablation Study on Main Components}. The table shows performance scores across different evaluation metrics for various model configurations on MULTIPLE dataset.}
\centering
\resizebox{\linewidth}{!}{
\begin{tabular}{l|c|ccc|c}
\toprule
\textbf{Method}  &\textbf{VACoT} & \textbf{Character} & \textbf{Object} & \textbf{Char. + Obj.} & \textbf{Average↑} \\
\midrule
BAGEL  && 5.17& 6.64& 6.24 & 6.02 \\
BAGEL  &\checkmark& 7.17 &8.65 &7.86 &7.89 \\
Ours w/o SFT  &\checkmark&7.61 &9.00 &7.56 &8.06 \\
Ours w/o GRPO  &\checkmark& 7.12& 9.07& 8.20&8.13 \\
\rowcolor{gray!20}
\textbf{Ours}  &\checkmark& 7.82& 9.21& 8.30& 8.44\\
\bottomrule
\end{tabular}
}
\label{tab:ablation_main}
\end{table}

\noindent\textbf{Analysis of Planning and Refinement Components.}
To further understand the contribution of our core components, we conduct ablation studies on Visual Adaptive Planning and Iterative Refinement, with results in Table~\ref{tab:ablation_components}. Removing Visual Adaptive Planning (L1) decreases performance across all metrics, demonstrating that explicitly designating the checking focus enables higher-quality evaluation and more valuable editing instructions. Similarly, removing Iterative Refinement (L2) leads to performance drops, indicating that the reasoning and refinement process provides proper direction toward better results.

\begin{table}[t!]
\caption{\textbf{Ablation Study on Planning and Refinement Components}. The table shows performance scores across different evaluation metrics for various planning configurations.}
\centering
\resizebox{\linewidth}{!}{
\begin{tabular}{l|ccc|c}
\toprule
\textbf{Method} & \textbf{Character} & \textbf{Object} & \textbf{Char. + Obj.} & \textbf{Average↑} \\
\midrule
Ours w/o Visual Adaptive Planning & 6.98& 8.81& 7.97& 7.92\\
Ours w/o Iterative Refinement & 7.20& 9.28& 8.18& 8.22\\
\rowcolor{gray!20}
\textbf{Ours} & 7.82& 9.21& 8.30& 8.44 \\
\bottomrule
\end{tabular}
}
\label{tab:ablation_components}
\end{table}

\noindent\textbf{Effect of Maximum Iterations.}
To investigate how many iterations the model maximally needs to refine an image, we conduct experiments on the multiple characters setting of the OmniContext benchmark. The results are shown in Table~\ref{tab:ablation_iterations}. The PF-score means the extent of prompt following, while the SC-score means subject consistency. The results demonstrate that iterative refinement achieves significant improvements compared to single-pass generation. However, images that can be successfully corrected are typically refined within 3 iterations. Increasing the maximum iteration count beyond this point only contributes to the challenging cases that the model struggles to resolve, where additional iterations may actually lead to degraded performance as the model over-corrects or introduces new errors. \textbf{More iterative refinement results can be found in Supp.}

\begin{table}[t!]
\caption{\textbf{Ablation Study on Maximum Iterations}. Experiments are conducted on Multiple Characters setting of OmniContext.}
\centering

\resizebox{\linewidth}{!}{
\begin{tabular}{c|ccc}
\toprule
\textbf{Maximum Iterations} & \textbf{PF-Score} & \textbf{SC-Score} &  \textbf{Average-Score↑} \\
\midrule
1 & 8.02& 6.96& 7.20\\
2 & 8.54& 6.88& 7.42\\
3 & 8.98& 7.10& 7.82\\ 
5 & 8.96& 7.04& 7.70\\ 

\bottomrule
\end{tabular}
}
\label{tab:ablation_iterations}
\end{table}

%% file: sec/X_suppl.tex
\clearpage
\setcounter{page}{1}
\maketitlesupplementary
\noindent The supplementary material is organized as follows:
\begin{itemize}
    \item \textbf{Section~\ref{sec:implement}} introduces implementation details, including the construction of planning and correction datasets, system prompts, and training configurations.
    \item \textbf{Section~\ref{sec:supp_exp}} provides comprehensive experimental results, including more complex multi-reference generation experiments with style reference, and additional ablation studies on reward model design.
    \item \textbf{Section~\ref{sec:supp_qual}} demonstrates qualitative results, showing more iterative refinement examples, detailed comparisons with baseline methods, reasoning process comparisons with text-based CoT methods, and analysis of failure cases.
\end{itemize}

\section{Implementation Details}
\label{sec:implement}

\subsection{Dataset Construction}

We construct our planning and correction datasets using a systematic approach based on multi-reference image editing scenarios. Our dataset construction pipeline consists of two main components: planning dataset generation and correction dataset generation.

\subsubsection{Data Source and Sampling}

We randomly sample 4,000 multi-reference data samples from the Echo-4o dataset~\cite{ye2025echo}. Each sample contains a triplet of {reference images, editing instruction, ground truth image}, where reference images provide visual context, editing instruction describes the desired editing operation in natural language, and ground truth image represents the expected output after applying the editing instruction.

\subsubsection{Planning Dataset Generation}

The planning dataset aims to teach the model to decompose complex editing instructions into structured, actionable visual plans. For each sample triplet \{reference images, instruction, ground truth image\}, we generate an adaptive visual plan that serves as a step-by-step checklist for the editing process.

\begin{figure}[!t]
\centering
\includegraphics[width=1\linewidth]{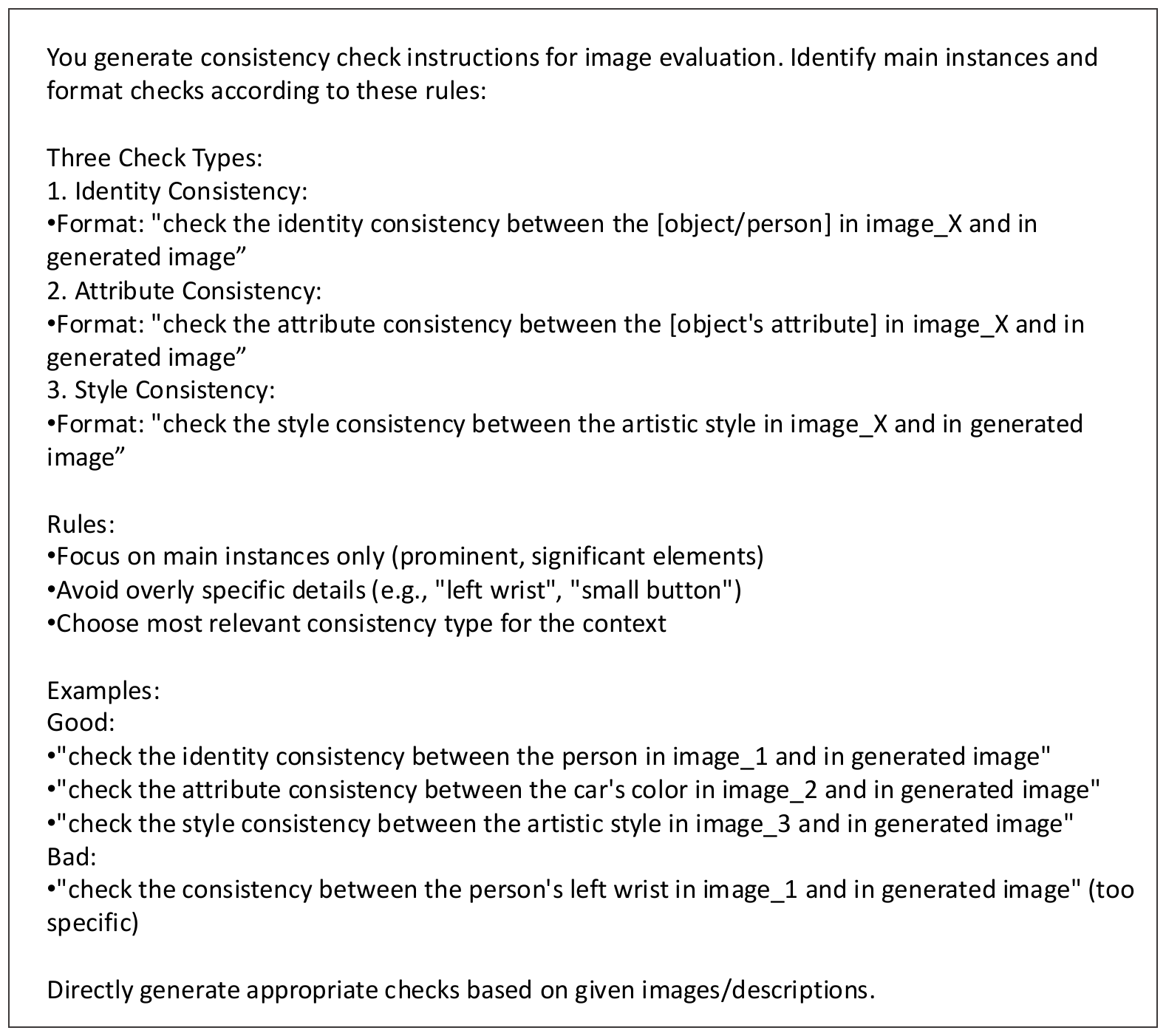}
\caption{\textbf{System prompt for visual plan generation.} The prompt guides the model to create structured, actionable plans for image editing tasks.}
\label{fig:planning_prompt}
\end{figure}

We employ Gemini-2.5-Flash~\cite{google_gemini2_flash} as our plan generation model, using the system prompt detailed in Figure~\ref{fig:planning_prompt}. The generated visual plans contain key visual elements to identify from reference images, specific editing operations to perform, and quality criteria for evaluating the editing results.

\subsubsection{Correction Dataset Generation}

The correction dataset focuses on training the model's self-evaluation and iterative refinement capabilities. For each sample, we construct a 5-tuple \{reference images, instruction, ground truth image, visual plan, current generated image\} where the current generated image is generated from a sub-optimal model with the \{reference images, editing instruction, ground truth image\}.

\begin{figure}[!t]
\centering
\includegraphics[width=1\linewidth]{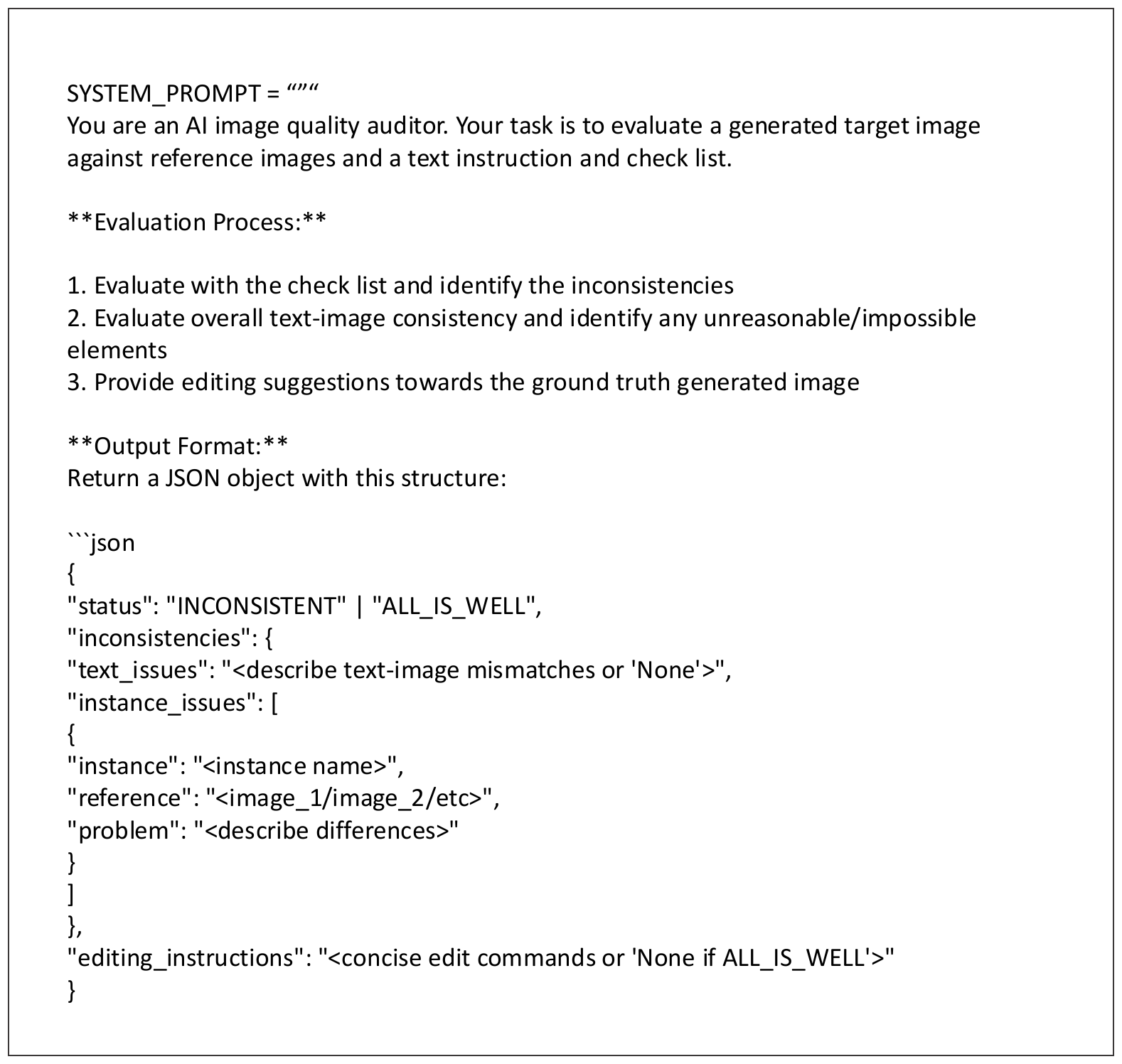}
\caption{\textbf{System prompt for evaluation and correction generation.} The prompt enables the model to assess current results and provide specific editing instructions for improvement.}
\label{fig:eval_prompt}
\end{figure}

Using this 5-tuple as input, we employ the system prompt shown in Figure~\ref{fig:eval_prompt} to generate self-evaluation results that provide detailed assessment of the current image quality against the visual plan, and editing instructions that offer specific guidance for improving the current image to better match the target. 

\subsection{Training Sequence}

The Bagel team's implementation enables flexible formatting of interleaved image-text data for training through the \texttt{add\_text} and \texttt{add\_image} functions. A critical parameter is \texttt{need\_loss}, which indicates whether ground truth is expected for loss calculation during training.

\begin{figure*}[!t]
\centering
\includegraphics[width=1\linewidth]{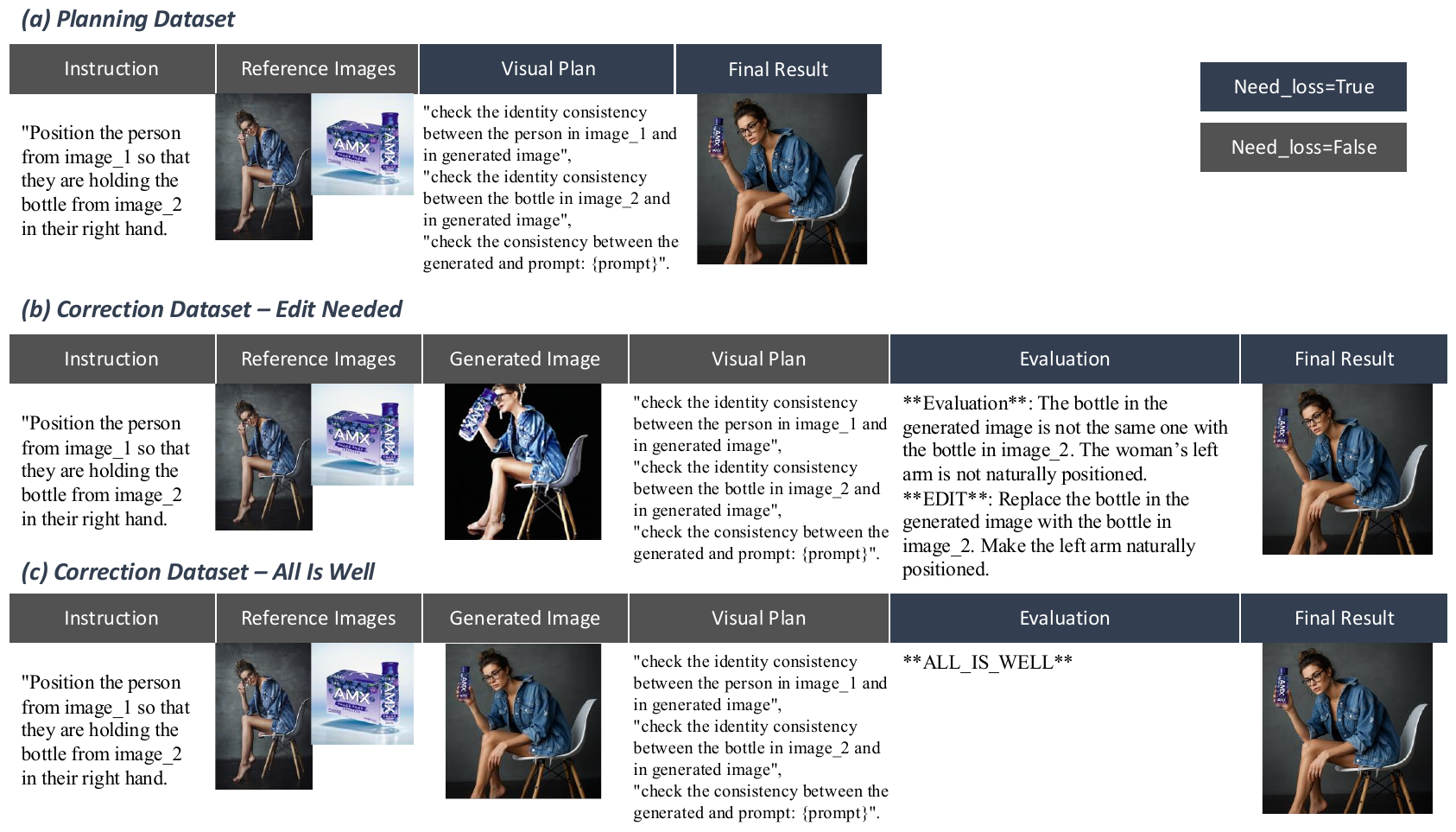}
\caption{\textbf{Training sequence design for different data types.} Gray blocks indicate no loss calculation, while blue blocks require loss calculation.}
\label{fig:data_sequence}
\end{figure*}

We demonstrate our training sequence design in Figure~\ref{fig:data_sequence}, where gray blocks represent segments without loss calculation (\texttt{need\_loss=False}) and blue blocks represent segments requiring loss calculation (\texttt{need\_loss=True}). Our training data encompasses three distinct types of samples: planning data samples, correction data samples for sub-optimal results, and correction data samples for perfect results.

In the planning data samples, the model learns to generate structured visual plans given reference images and instructions. For correction data samples with sub-optimal results, the model practices identifying discrepancies and providing specific improvement guidance. For correction data samples with perfect results, the model learns to recognize when no further editing is needed and provide appropriate feedback.

\subsection{Object Similarity Score}
\begin{figure}[!t]
\centering
\includegraphics[width=1\linewidth]{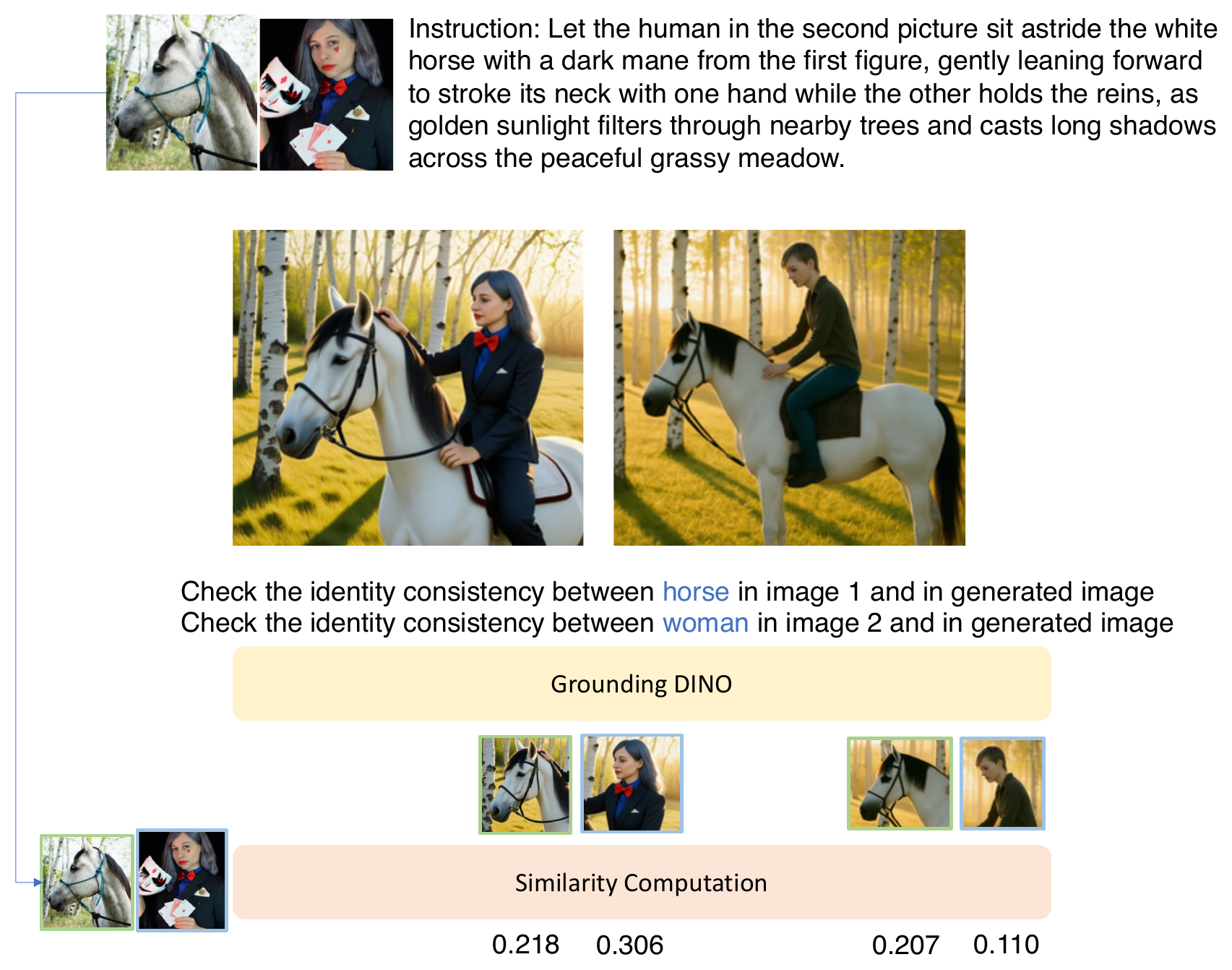}
\caption{\textbf{The object similarity score used in GRPO training.}}
\label{fig:objsim}
\end{figure}

As shown in Figure~\ref{fig:objsim}, we compute object similarity scores to measure identity consistency between reference and generated images. We first parse the visual plan checklist to identify target objects. For example, the checklist specifies: ``Check the identity consistency between horse in image 1 and in generated image" and ``Check the identity consistency between woman in image 2 and in generated image."

We use GroundingDINO to locate these objects in both reference and generated images based on the text descriptions. After extracting bounding boxes for the specified objects, we compute DINO feature similarity between the cropped regions. Higher scores indicate better object identity consistency. This object similarity score provides fine-grained feedback for GRPO training on whether the generated image preserves the visual identity of key objects from the reference images.

To validate whether the object similarity score is reasonable, we test it on our training dataset with both sub-optimal generation results and ground truth images. We find that in 78.3\% of the data samples, the ground truth has a higher score than the sub-optimal generation, which means our object similarity score effectively distinguishes between high-quality and low-quality results in most cases.

\subsection{Training Configuration}

We train VACoT on 8 NVIDIA A800 GPUs using the Adam optimizer with a constant learning rate of 2e-5. We apply a linear warmup schedule that gradually increases the learning rate from zero over the first 500 steps. During training, we randomly sample from our planning and correction datasets, packing sequences to a maximum length of 32,000 tokens per batch. All model parameters except ViT and VAE are jointly optimized throughout the training process.

\section{Experiments}
\label{sec:supp_exp}

\subsection{Complete Comparison with other CoT methods}
The evaluation of the OmniContext benchmark consists of two dimensions: Prompt Following (PF score) and Subject Consistency (SC score). Due to space constraints in the main paper, we present the detailed scores here. As shown in Table~\ref{tab:supp_main}, our method consistently outperforms other CoT methods on subject consistency, demonstrating that our design and training approach effectively enhances visual-aware generation capabilities.

\begin{table*}[!t]
\centering
\caption{\textbf{Comparison of different multi-modal Chain-of-Thought on OmniContext.}}
\resizebox{\textwidth}{!}{%
\begin{tabular}{l|cc|cc|cc|cc|cc|cc|c}
\toprule
& \multicolumn{2}{c|}{\textbf{Multi Character}} & \multicolumn{2}{c|}{\textbf{Multi Object}} & \multicolumn{2}{c|}{\textbf{Multi Char Obj}} & \multicolumn{2}{c|}{\textbf{Scene Character}} & \multicolumn{2}{c|}{\textbf{Scene Object}} & \multicolumn{2}{c|}{\textbf{Scene Char Obj}} & \textbf{Overall} \\
\cmidrule(lr){2-3} \cmidrule(lr){4-5} \cmidrule(lr){6-7} \cmidrule(lr){8-9} \cmidrule(lr){10-11} \cmidrule(lr){12-13}
\textbf{Model} & \textbf{PF} & \textbf{SC} & \textbf{PF} & \textbf{SC} & \textbf{PF} & \textbf{SC} & \textbf{PF} & \textbf{SC} & \textbf{PF} & \textbf{SC} & \textbf{PF} & \textbf{SC} & \textbf{Average} \\
\midrule
UiG & 6.90 & 4.92 & 8.24 & 8.72 & 8.30 & 8.30 & 5.54 & 5.20 & 6.94 & 7.60 & 7.08 & 7.38 & 6.986 \\
UniCoT & 8.34 & 6.48 & 8.92 & 9.06 & 8.34 & 7.96 & 7.74 & 6.82 & 8.12 & 8.38 & 8.34 & 8.10 & 7.893 \\
Ours & 8.98 & 7.10 & 9.26 & 9.26 & 8.60 & 8.18 & 8.27 & 7.24 & 8.80 & 8.60 & 8.10 & 8.16 & 8.257 \\
\bottomrule
\end{tabular}%
}
\label{tab:supp_main}
\end{table*}

\subsection{Ablation Study: Reward Model Design}
\paragraph{Reward Function Analysis}
Table~\ref{tab:ablation_reward} presents an ablation study on different reward function configurations for GRPO training. We compare using object similarity score (ObjSimScore) alone, ObjSimScore combined with CLIP Score, and the full combination of ObjSimScore, PickScore, and CLIPScore. The results show that adding CLIPScore to ObjSimScore improves performance across all categories, achieving the best average score of 8.44. However, incorporating PickScore degrades performance, suggesting it may introduce conflicting optimization signals. The optimal reward function combines ObjSimScore and CLIPScore, balancing object-specific similarity with semantic alignment.

\begin{table}[t!]
\caption{\textbf{Ablation Study on GRPO Reward Functions}. The table shows performance scores across different reward function configurations.}
\centering
\resizebox{\linewidth}{!}{
\begin{tabular}{l|ccc|c}
\toprule
\textbf{Reward Function} & \textbf{Character} & \textbf{Object} & \textbf{Char. + Obj.} & \textbf{Average↑} \\
\midrule
ObjSimScore& 7.45& 8.92& 7.88& 8.08\\
ObjSimScore + CLIPScore& 7.82& 9.21& 8.30& 8.44\\
ObjSimScore + PickScore + CLIPScore& 6.89& 8.15& 7.42& 7.49\\
\bottomrule
\end{tabular}
}
\label{tab:ablation_reward}
\end{table}

\subsection{Complex Multi-reference Generation}

\begin{figure*}[!t]
\centering
\includegraphics[width=1\linewidth]{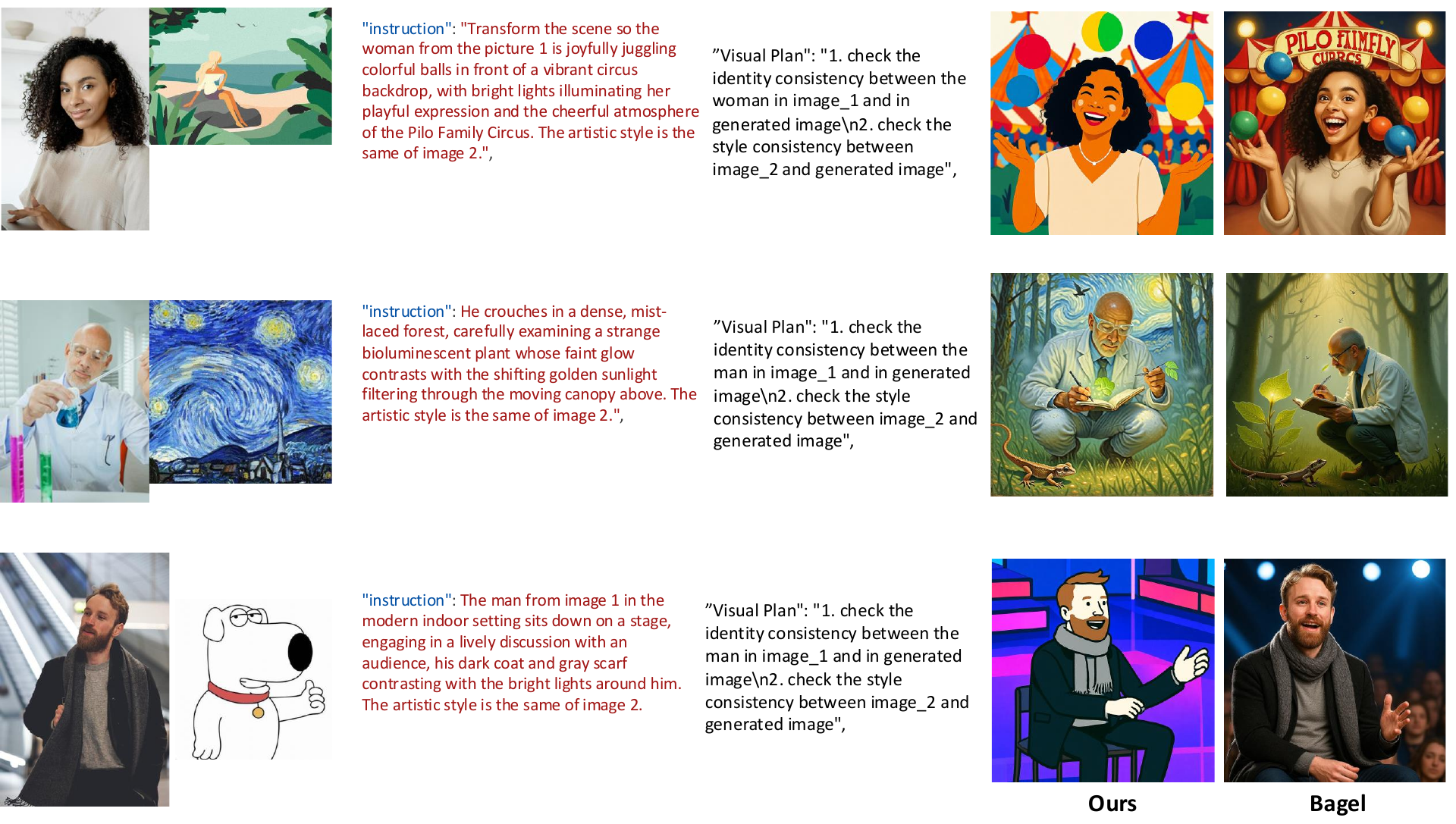}
\caption{\textbf{Results on complex multi-reference generation.}}
\label{fig:style}
\end{figure*}

We evaluate the zero-shot capability of our method on complex multi-reference generation tasks that require both identity preservation and style transfer. As illustrated in Figure~\ref{fig:style}, the baseline BAGEL model fails to generate images that accurately reflect the target style while maintaining the subject's identity. In contrast, our approach demonstrates robust performance on this challenging task despite not being trained on style transfer data. This zero-shot capability highlights the effectiveness of our method in handling complex multi-reference scenarios where multiple visual attributes must be simultaneously controlled.

\section{More Qualitative Results}
\label{sec:supp_qual}

\subsection{Iterative Refinement Examples}

\begin{figure*}[!t]
\centering
\includegraphics[width=1\linewidth]{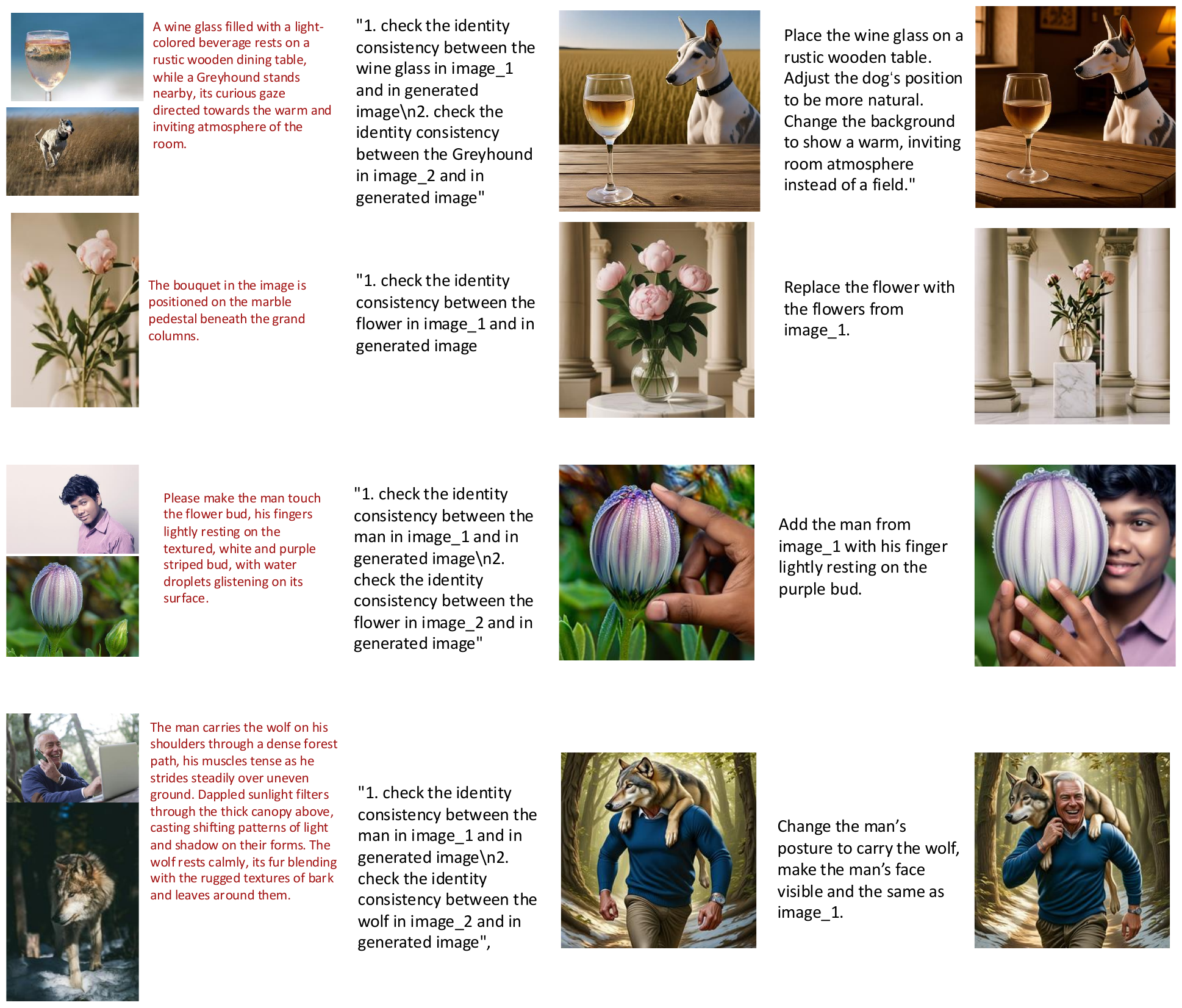}
\caption{\textbf{Iterative refinement during generation.} }
\label{fig:iterative_supp}
\end{figure*}

We demonstrate the effectiveness of our iterative refinement mechanism through several representative examples. As shown in Figure~\ref{fig:iterative_supp}, the initially generated images often contain various types of defects, including identity distortion, missing identity features, or unreasonable elements. Our method systematically identifies these issues and generates appropriate editing instructions to progressively correct them. Through this iterative process, the model can automatically refine the generated images to achieve higher quality results that better preserve the target identity while maintaining visual coherence.

\subsection{Comprehensive Method Comparisons}

\begin{figure*}[!t]
\centering
\includegraphics[width=1\linewidth]{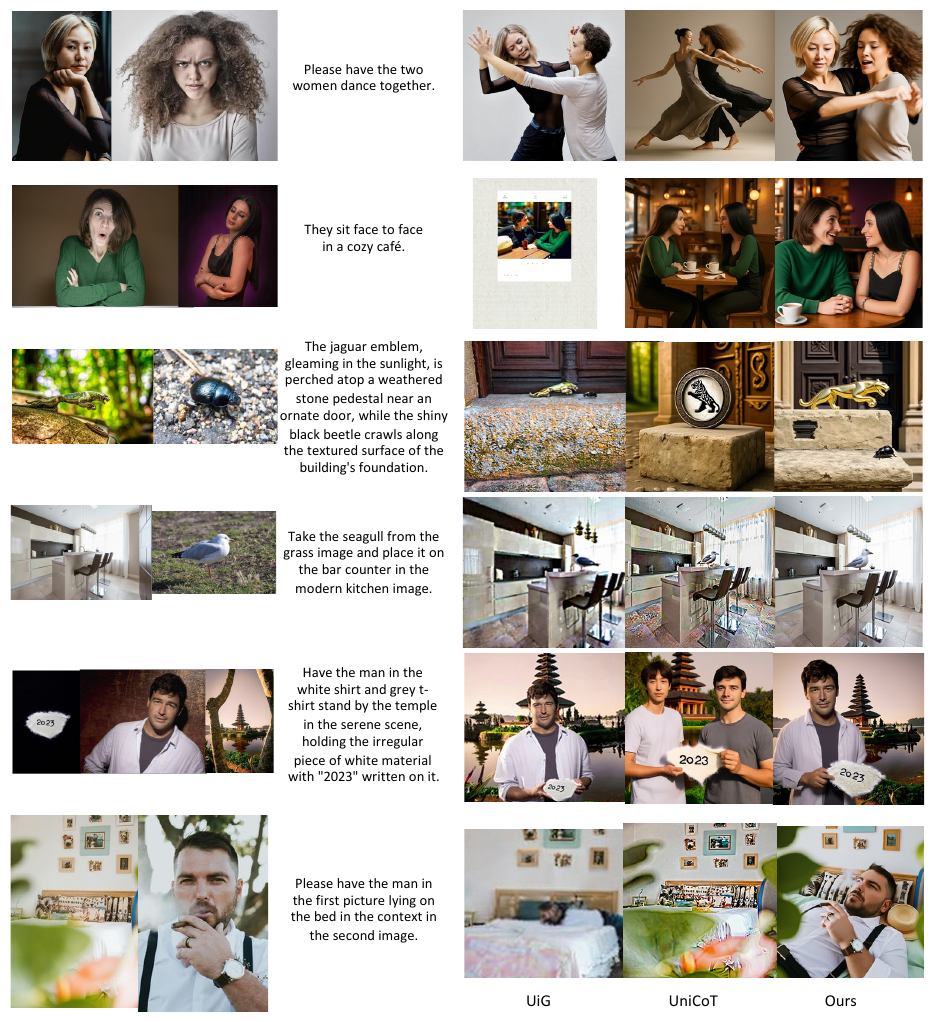}
\caption{\textbf{Qualitative comparison with other CoT-based methods on OmniContext.} Our method demonstrates superior identity preservation and subject consistency across diverse generation scenarios compared to existing approaches.}
\label{fig:qual_supp}
\end{figure*}

We present extensive qualitative comparisons with other Chain-of-Thought (CoT) based methods to demonstrate the superiority of our approach. As illustrated in Figure~\ref{fig:qual_supp}, our method consistently achieves higher subject consistency and superior identity preservation compared to existing CoT-based approaches. The visual results clearly show that our method maintains better facial features, expressions, and overall identity characteristics while generating contextually appropriate images. 

\subsection{Ablation Study on Training Stages}

\begin{figure*}[!t]
\centering
\includegraphics[width=1\linewidth]{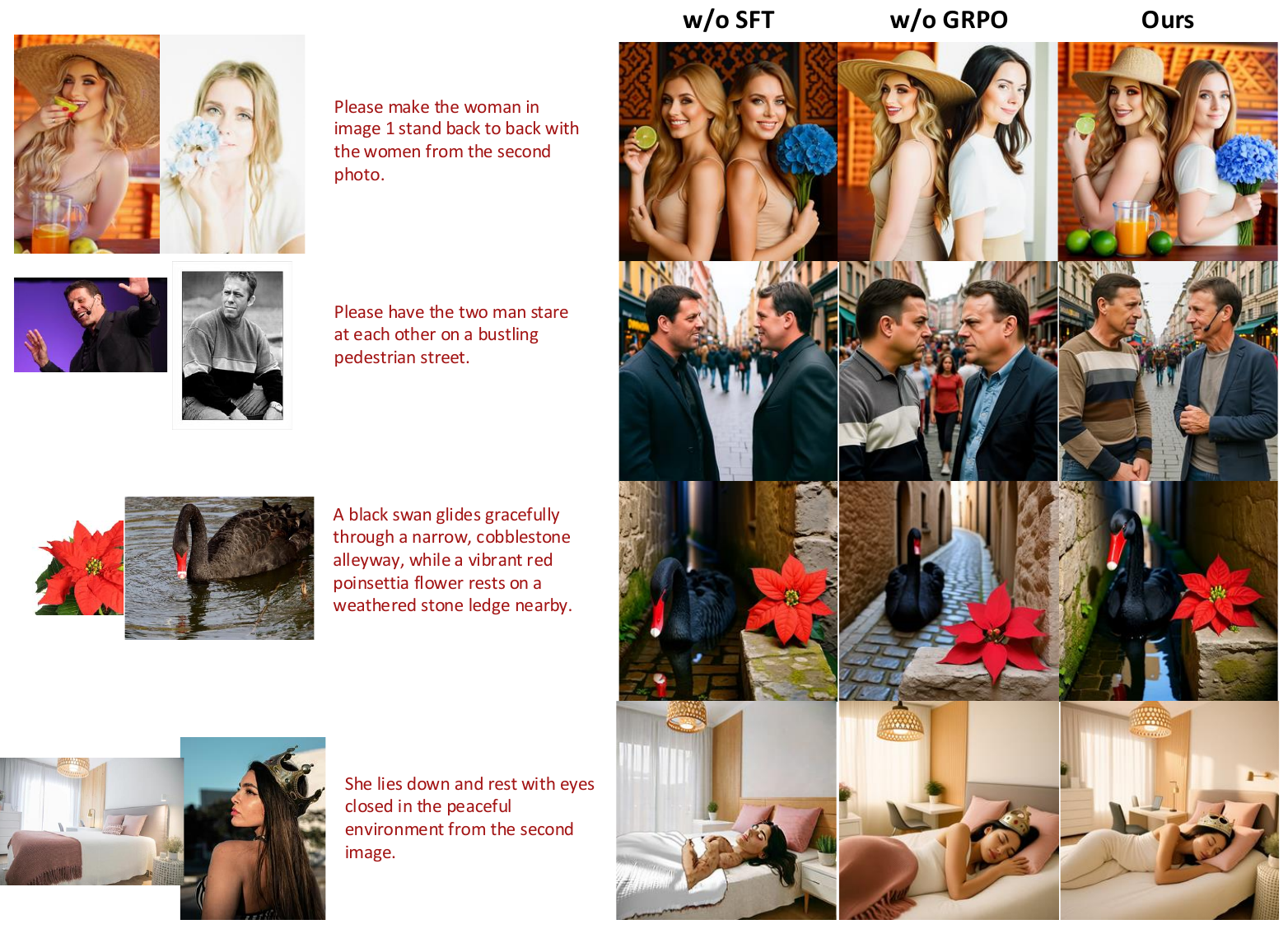}
\caption{\textbf{Ablation study on training stages.} Comparison of our full method against variants without SFT and without GRPO, showing the relative importance of each training component.}
\label{fig:ablation_supp}
\end{figure*}

We conduct a qualitative ablation study to analyze the contribution of different training stages in our method. Specifically, we evaluate three configurations: our full method, a variant without Supervised Fine-Tuning (SFT), and a variant without GRPO. The results are presented in Figure~\ref{fig:ablation_supp}.

The results reveal that both training stages contribute to the overall performance, but with different levels of importance. SFT plays a more critical role in establishing the foundational capabilities of the model, as evidenced by the significant performance degradation when it is removed. GRPO provides additional refinement and optimization. The combination of both training stages achieves the best performance, demonstrating that our two-stage training strategy effectively enhances the visual-aware ability of the model.

\subsection{Reasoning Process Comparison}
\begin{figure*}[!t]
\centering
\includegraphics[width=1\linewidth]{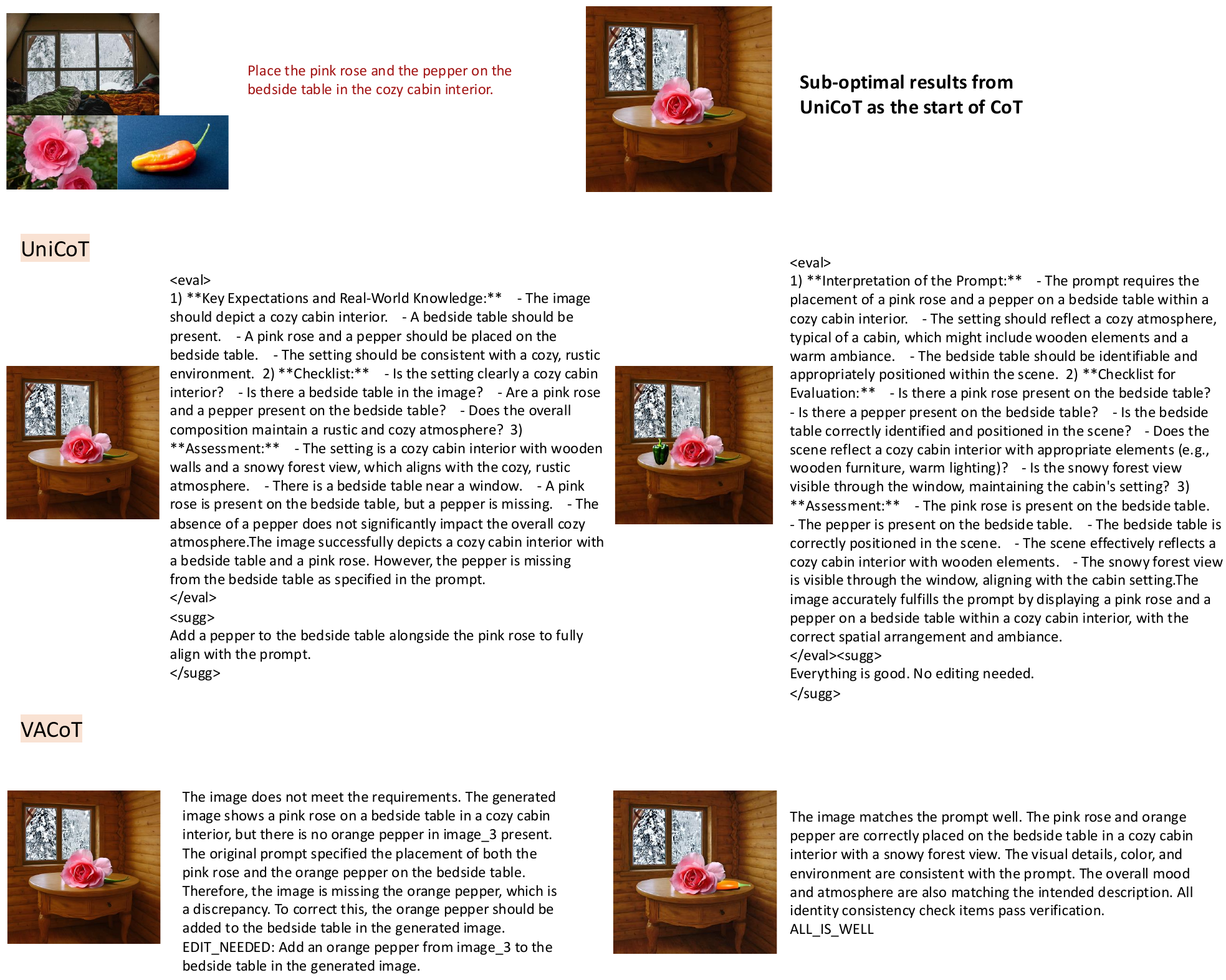}
\caption{\textbf{Reasoning comparison with UniCoT.}}
\label{fig:comp_w_unicot}
\end{figure*}

We conduct a detailed comparison of our reasoning process against UniCoT to highlight the advantages of our approach. Starting with a sub-optimal initial result generated by UniCoT, we demonstrate the self-evaluation capabilities of both methods in Figure~\ref{fig:comp_w_unicot}. UniCoT's reasoning process focuses on text alignment with the user prompt, checking only for the presence of specified objects (e.g., whether a pepper exists in the image) while overlooking critical visual attributes such as identity-specific features. In the shown example, UniCoT correctly identifies the presence of a pepper but fails to recognize that the pepper's color does not match the specified requirements, thus missing a fundamental aspect of the prompt. In contrast, our method employs visual-aware reasoning that can accurately identify such discrepancies and provide appropriate corrections. 

\subsection{Failure Case Analysis}

We find that problems that can be fixed are usually resolved within 2 iterations. Beyond this point, the remaining issues become much harder to correct, and additional iterations often make things worse.

Our analysis shows two main reasons why more iterations fail. First, repeated editing degrades image quality by accumulating noise and artifacts with each modification. Second, the problems that persist after 2 iterations are typically fundamental issues that our method struggles to identify correctly or address effectively. When the model tries to fix these harder problems, it often misdiagnoses the issue or applies inappropriate corrections.

\begin{figure*}[!t]
\centering
\includegraphics[width=1\linewidth]{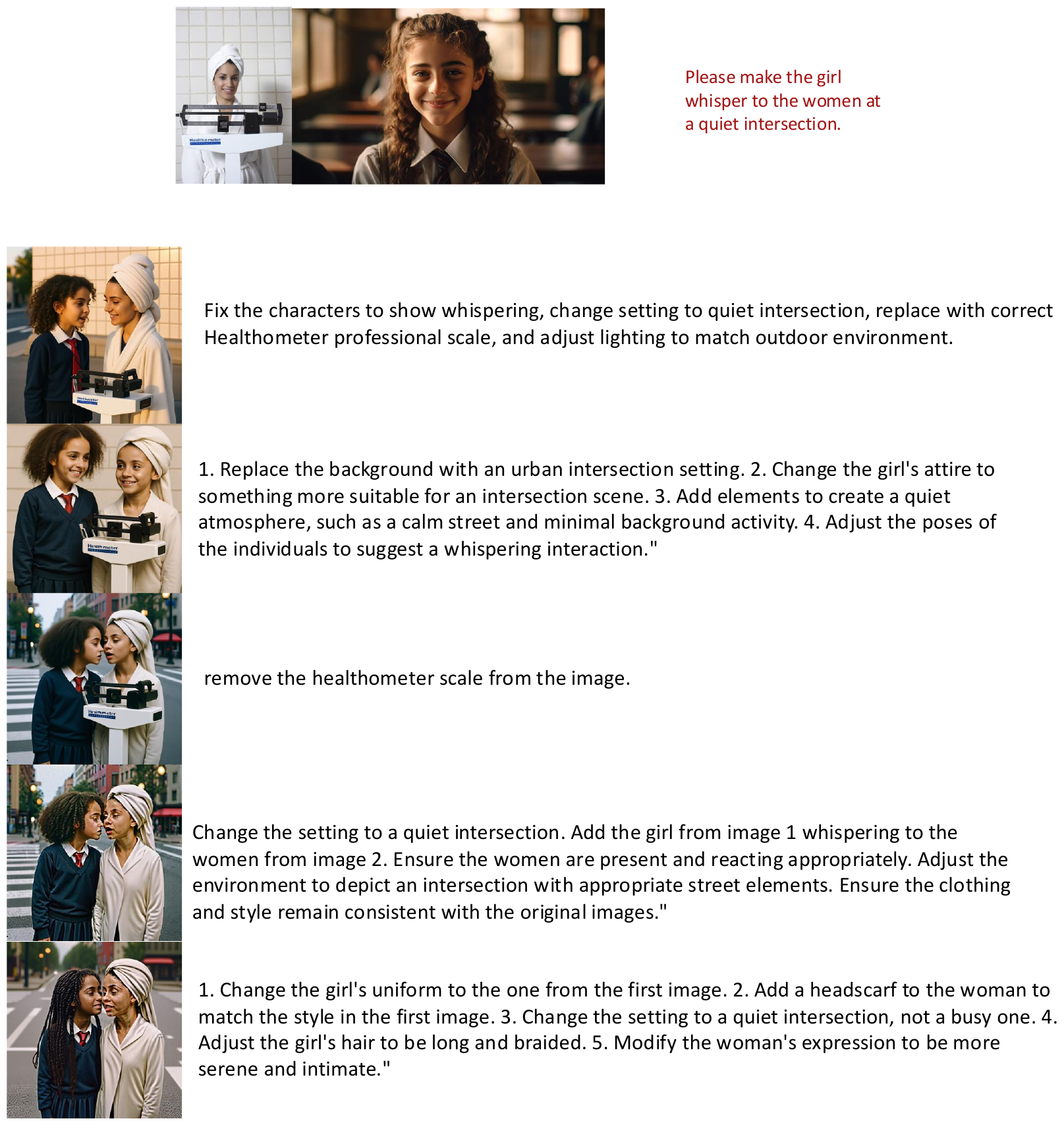}
\caption{\textbf{Failure case analysis.}}
\label{fig:failure}
\end{figure*}

Figure~\ref{fig:failure} demonstrates these failure cases. During the 5 iterations shown, each iteration identifies a new problem, but these problems are not important or meaningful. Instead of improving the result, attempting to fix these minor issues makes the final output worse.

